\documentclass[10pt,twocolumn,letterpaper]{article}

\usepackage{cvpr}              

\usepackage[dvipsnames]{xcolor}

\usepackage{graphicx}
\usepackage{amsmath}
\usepackage{amssymb}
\usepackage{booktabs}

\usepackage{times}
\usepackage{epsfig}

\usepackage{verbatim}

\usepackage{algorithm}
\usepackage{algorithmicx}
\usepackage{algpseudocode}

\usepackage{color}

\usepackage{ctable}
\usepackage{makecell}
\usepackage{tabularx}
\usepackage{multirow}
\usepackage{multicol}
\usepackage{arydshln}

\usepackage{pifont}
\definecolor{cvprblue}{rgb}{0.21,0.49,0.74}
\usepackage[pagebackref,breaklinks,colorlinks,citecolor=cvprblue]{hyperref}

\usepackage[capitalize]{cleveref}
\crefname{section}{Sec.}{Secs.}
\Crefname{section}{Section}{Sections}
\Crefname{table}{Table}{Tables}
\crefname{table}{Tab.}{Tabs.}

\usepackage[accsupp]{axessibility}

\def\paperID{7178} 
\def\confName{CVPR}
\def\confYear{2024}

\begin{document}

\title{Joint Reconstruction of 3D Human and Object \\ via Contact-Based Refinement Transformer}

\author{
  Hyeongjin Nam$^{1,3*}$ \hskip1.6em Daniel Sungho Jung$^{2,3*}$ \hskip1.6em Gyeongsik Moon$^{4}$ \hskip1.6em Kyoung Mu Lee$^{1,2,3}$ \\
   $^{1}$Dept. of ECE\&ASRI, $^{2}$IPAI, Seoul National University, Korea  \\ 
   $^{3}$SNU-LG AI Research Center,
   $^{4}$Codec Avatars Lab, Meta \\
   {\tt\small \{namhjsnu28, dqj5182\}@snu.ac.kr, mks0601@meta.com, kyoungmu@snu.ac.kr} 
}

\maketitle

\begin{abstract}
Human-object contact serves as a strong cue to understand how humans physically interact with objects.
Nevertheless, it is not widely explored to utilize human-object contact information for the joint reconstruction of 3D human and object from a single image.
In this work, we present a novel joint 3D human-object reconstruction method (CONTHO) that effectively exploits contact information between humans and objects.
There are two core designs in our system: 1) 3D-guided contact estimation and 2) contact-based 3D human and object refinement.
First, for accurate human-object contact estimation, CONTHO initially reconstructs 3D humans and objects and utilizes them as explicit 3D guidance for contact estimation. 
Second, to refine the initial reconstructions of 3D human and object, we propose a novel contact-based refinement Transformer that effectively aggregates human features and object features based on the estimated human-object contact.
The proposed contact-based refinement prevents the learning of erroneous correlation between human and object, which enables accurate 3D reconstruction.
As a result, our CONTHO achieves state-of-the-art performance in both human-object contact estimation and joint reconstruction of 3D human and object.
The code is publicly available\footnote{\url{https://github.com/dqj5182/CONTHO_RELEASE}}.
\end{abstract}

\renewcommand{\thefootnote}{\fnsymbol{footnote}}
\footnotetext[1]{equal contribution.}
\renewcommand{\thefootnote}{\arabic{footnote}}
\section{Introduction}
Joint reconstruction of 3D human and object is an essential task for various applications of immersive experiences of AR/VR and robot manipulation of robotics.
In essence, the task aims to learn a meaningful human-object interaction that further improves the reconstruction of humans and objects.
Physical human-object contact is notably one of the most prevalent and basic interactions that humans make with objects.

\begin{figure}[t]
\begin{center}
\includegraphics[width=1.0\linewidth]{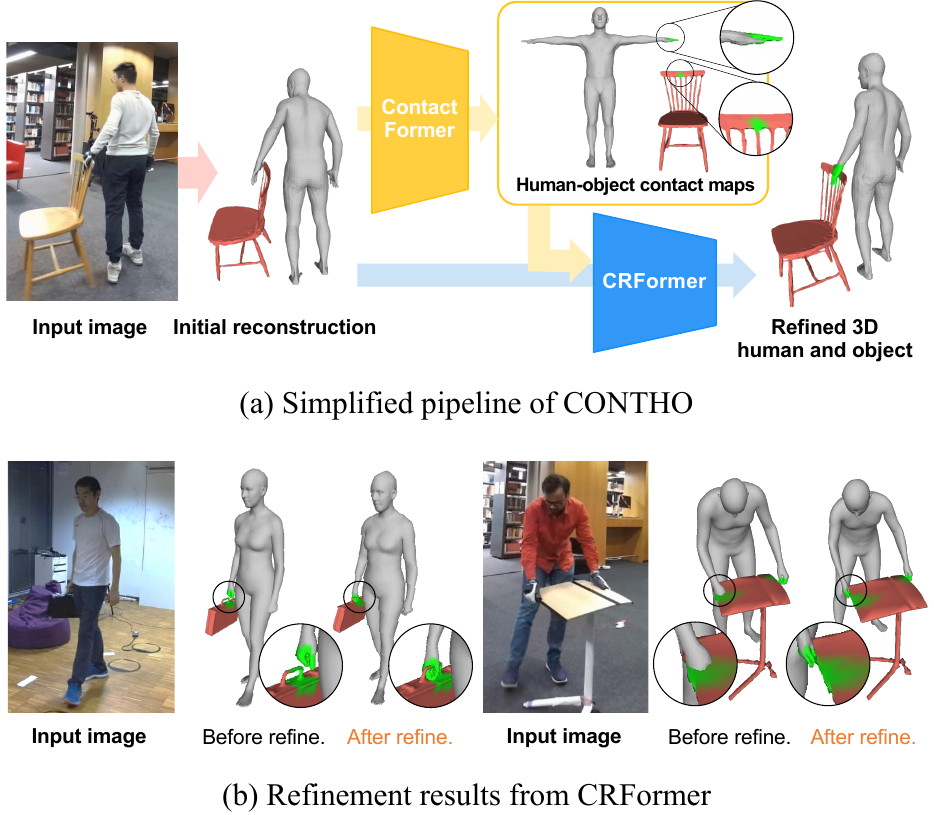}
\end{center}
\vspace*{-4.2mm}
\caption{
\textbf{Overview of CONTHO.}
Our proposed CONTHO estimates human-object contact maps through our proposed \textcolor[RGB]{230,170,10}{\textbf{ContactFormer}} and exploits the contact maps for 3D human and object refinement with \textcolor[RGB]{0,130,245}{\textbf{CRFormer}}.
The green color indicates human-object contact regions estimated from \textcolor[RGB]{230,170,10}{\textbf{ContactFormer}}.
}
\label{fig:introduction}
\vspace*{-2.0mm}
\end{figure}

Although human-object contact is a strong cue in joint reconstruction of 3D human and object, recent works of human-object interaction have been studied separately in two major tracks: 1) human-object contact estimation and 2) 3D human and object reconstruction.
The recent research track for human-object contact estimation~\cite{huang2022capturing,tripathi2023deco,shimada2022hulc} predicts a contact map on the surface of a pre-defined human body model~\cite{loper2015smpl,pavlakos2019expressive} without reconstructing 3D human and object.
Another research track for 3D human and object reconstruction~\cite{zhang2020perceiving,xu2021d3d,xie2022chore,xie2023visibility} does not yet sufficiently explore how to extract and utilize contact information for the reconstruction.
For example, PHOSA~\cite{zhang2020perceiving} and D3D-HOI~\cite{xu2021d3d} heuristically pre-define contacting regions and follow the pre-defined regions as a hard constraint during their optimizations.
The pre-defined contacting regions can be different from the authentic ones in the image, which results in incorrect reconstructions of 3D human and object.

In this work, we integrate the two separate tracks with one unified framework, \textbf{CONTHO}~(\textbf{CONT}act-based 3D \textbf{H}uman and \textbf{O}bject reconstruction) that estimates human-object contact maps and exploits the contact maps for 3D human and object reconstruction, as shown in Figure~\ref{fig:introduction}.
In CONTHO, there are two core stages: 1) 3D-guided contact estimation and 2) contact-based 3D human and object refinement.
In the first stage, our proposed contact estimation Transformer~(\textcolor[RGB]{230,170,10}{\textbf{ContactFormer}}) utilizes initially reconstructed 3D human and object meshes as 3D guidance on 3D positional relationships between human and object.
In inferring contact, 3D positions of human and object surfaces provide valuable information about which parts of the human interact with the object. 
However, previous contact estimation methods~\cite{huang2022capturing,tripathi2023deco,shimada2022hulc} do not infer 3D geometric information of human and object surfaces during their estimation pipeline.
Unlike these methods, our ContactFormer utilizes the 3D positions of human and object surfaces along with image evidence, enabling 3D geometric reasoning about the relationship between 3D human and object. 
In the end, 3D-guided contact estimation provides accurate human-object contact maps, which benefits the next stage, the contact-based 3D human and object refinement.

\begin{figure}[t!]
\begin{center}
\includegraphics[width=1.0\linewidth]{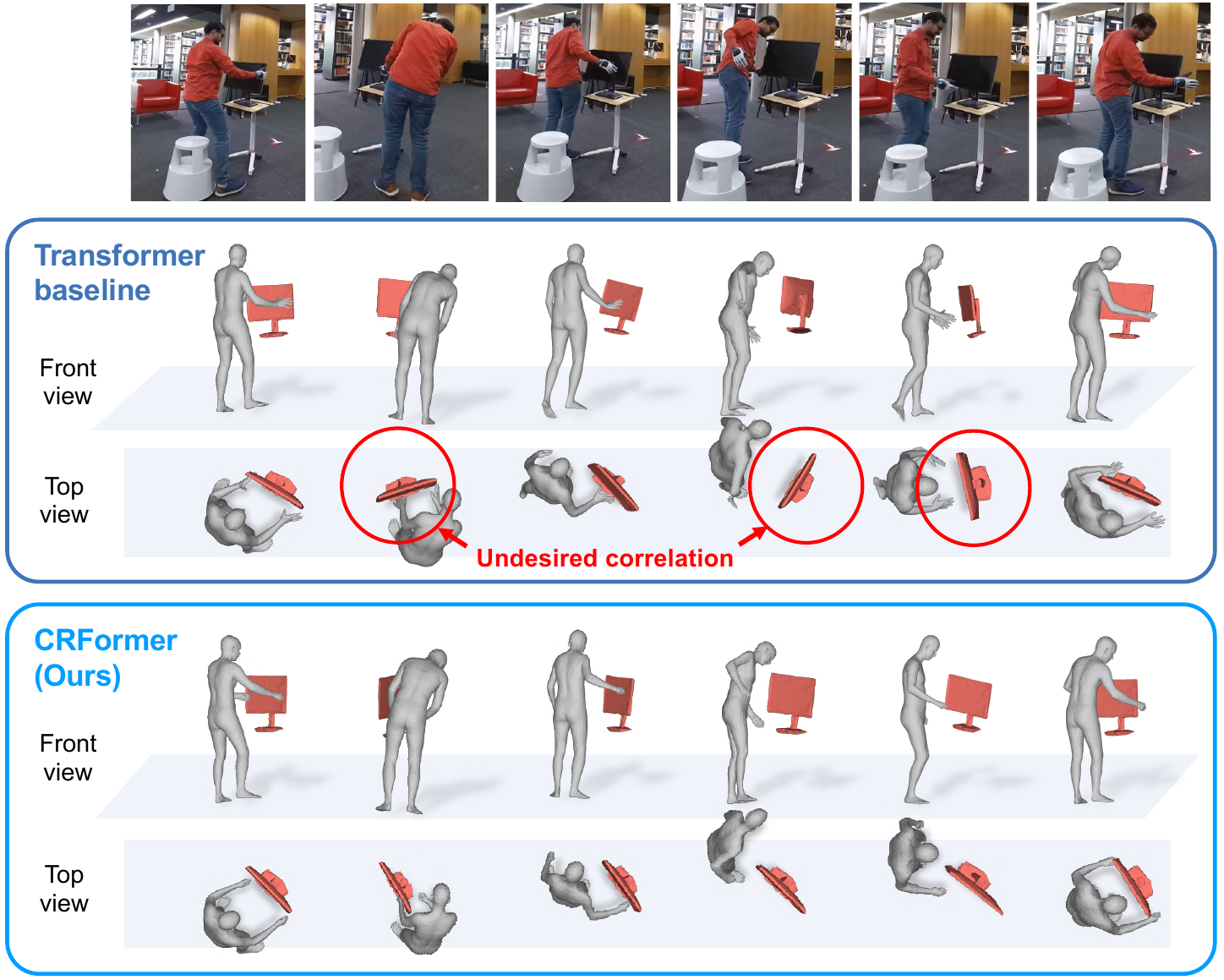}
\end{center}
\vspace*{-4.0mm}
\caption{
\textbf{Example of undesired human-object correlation.}
Due to the undesired human-object correlation in the Transformer baseline, the monitor display always faces toward the human head, which should not move as in the images.
Our proposed CRFormer effectively alleviates the undesired correlation, resulting in accurate reconstruction results.
}
\label{fig:undesired_bias}
\vspace*{-0.5mm}
\end{figure}

In the second stage, our proposed contact-based refinement Transformer~(\textcolor[RGB]{0,130,245}{\textbf{CRFormer}}) refines the initially reconstructed 3D human and object by effectively aggregating human and object features based on the estimated contact maps.
In the CRFormer, human and object features are selectively forwarded based on human-object contact maps to learn human-object interaction.
Such an approach has two advantages in 3D human and object refinement.
First, the CRFormer makes the human-object contact maps the main decisive signals that indicate which features to focus on.
While contact information is one of the most influential components for understanding relationships between humans and objects, human-object contact exists in small regions of the image and thus may often be neglected. 
Our CRFormer design explicitly spotlights the contact regions, making human-object contact a key signal for refinement.
Second, the CRFormer alleviates the undesired human-object correlation by removing features unrelated to physical interaction (\textit{i.e.}, contact).
A refinement network can learn undesired human-object correlation by easily capturing a strong bias of human pose and object pose, different from the actual appearance in the image.
When naively aggregating contact maps and image features~(Transformer baseline in Figure~\ref{fig:undesired_bias}), the refinement network reconstructs a monitor display always facing toward a human head, showing an undesired correlation between the object and the human.
On the other hand, our CRFormer considers human-object relations solely from the features of contacting regions based on human-object contact captured in the image, preventing undesired human-object correlation.
With these two strengths, our proposed CRFormer accurately refines 3D human and object with human-object contact maps.

As a result, we show that CONTHO achieves state-of-the-art performance in both human-object contact estimation and joint reconstruction of 3D human and object.
Our contributions can be summarized as follows.
\begin{itemize}
\item We propose CONTHO, which jointly reconstructs 3D human and object by exploiting human-object contact as a key signal in reconstruction.
\item To obtain precise human-object contact, we leverage intermediate 3D human and object reconstructions as explicit 3D guidance in contact estimation.
\item To accurately reconstruct 3D human and object, our proposed CRFormer effectively aggregates human features and object features based on contact information, while preventing learning undesired human-object correlation.
\item CONTHO largely outperforms previous methods in both human-object contact estimation and joint reconstruction of 3D human and object.
\end{itemize}

\section{Related works}
\noindent\textbf{Human-object contact estimation.}
Most of the pioneering works on human-object contact estimation represent contact in the form of 2D contact~\cite{chen2023detecting}, 3D joint-level contact~\cite{shimada2020physcap, rempe2021humor, rempe2020contact, zanfir2018monocular, zou2020reducing}, or 3D patch-level contact~\cite{muller2021self, fieraru2021learning, fieraru2020three}.
Recently, several works~\cite{hassan2021populating,huang2022capturing,tripathi2023deco,shimada2022hulc} tackle the problem of estimating a dense vertex-level contact map, defined on the human body surface~(\textit{i.e.}, SMPL~\cite{loper2015smpl}).
POSA~\cite{hassan2021populating} proposed a conditional variational autoencoder~(cVAE)~\cite{kingma2013auto} that outputs which vertices are likely to be in contact with objects, given a human pose without any use of image evidence.
BSTRO~\cite{huang2022capturing} demonstrated a Transformer that learns contextual relationships among body vertices.
DECO~\cite{tripathi2023deco} proposed a cross-attention-based network that jointly leverages human body parts and scene contexts for contact estimation.

These methods are simply trained with cross-entropy loss between the predicted and ground-truth (GT) contact maps, without learning 3D geometry of human and object surfaces.
On the other hand, our CONTHO jointly learns human-object contact maps along with reconstructing the 3D human and object meshes, which has two noticeable advantages in contact estimation.
First, 3D human and object meshes provide guidance on where to focus on local image regions related to the human and object, with 2D vertex coordinates obtained by projecting the 3D meshes onto the input image.
Second, the per-vertex 3D coordinates provide a 3D positional relationship between human and object surfaces that allows 3D geometric reasoning on contact between human and object.
Under these two advantages, our 3D-guided contact estimation is much more effective in capturing human-object contact than the previous methods.

\begin{figure*}[t]
\begin{center}
\includegraphics[width=1.0\linewidth]{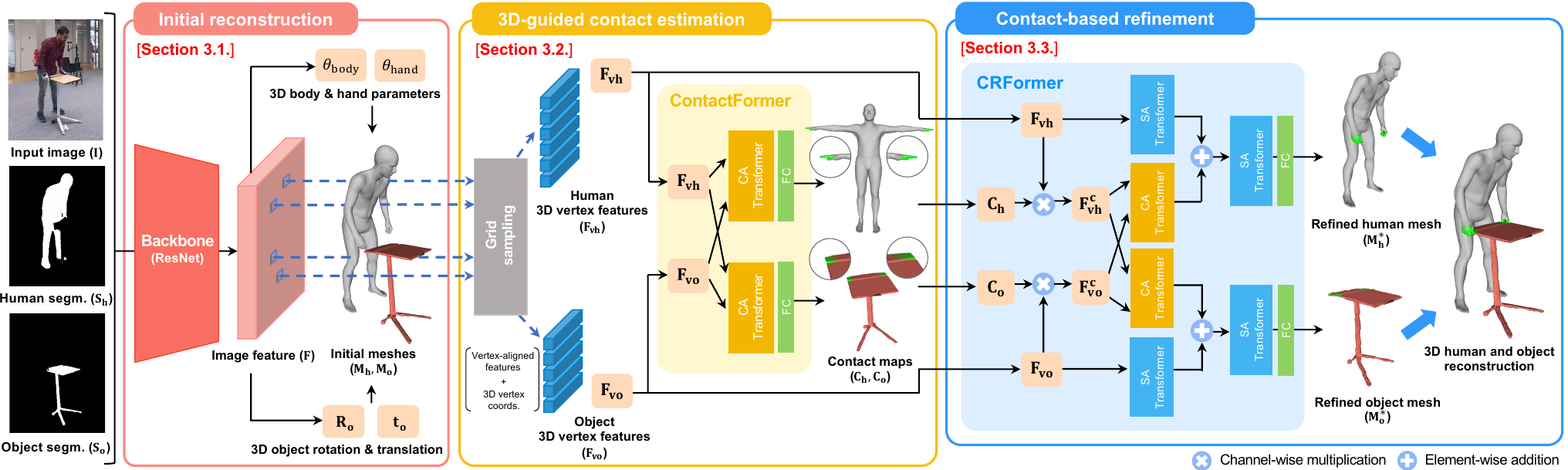}
\end{center}
\vspace*{-3.5mm}
\caption{
\textbf{Overall pipeline of CONTHO.}
Our method first reconstructs 3D human and object meshes ($\mathbf{M}_{\text{h}}$ and $\mathbf{M}_{\text{o}}$).
Then, the initial meshes are utilized to construct 3D vertex features ($\mathbf{F}_{\text{vh}}$ and $\mathbf{F}_{\text{vo}}$).
Subsequently, \textcolor[RGB]{230,170,10}{\textbf{ContactFormer}} estimates human-object contact maps ($\mathbf{C}_{\text{h}}$ and $\mathbf{C}_{\text{o}}$) from the 3D vertex features.
Lastly, \textcolor[RGB]{0,130,245}{\textbf{CRFormer}} aggregates the 3D vertex features based on the estimated contact maps to provide refined human and object meshes~($\mathbf{M}_{\text{h}}^{\ast}$ and $\mathbf{M}_{\text{o}}^{\ast}$).
The green color indicates the estimated contacting regions.
}
\label{fig:pipeline}
\end{figure*}

\noindent\textbf{3D human and object reconstruction.}
Most of the recent works~\cite{chen2019holistic++,zhang2020perceiving,xu2021d3d,xie2022chore,jiang2023full} of 3D human and object reconstruction are optimization-based approaches, which iteratively fit 3D human and object meshes to satisfy constraints of human-object interaction.
Holistic++~\cite{chen2019holistic++} designed human-object interaction priors based on human actions.
PHOSA~\cite{zhang2020perceiving} and D3D-HOI~\cite{xu2021d3d} each presented an optimization framework that fits human and object meshes based on pre-defined contact pairs to reason about human-object interaction.
CHORE~\cite{xie2022chore} proposed a two-stage approach, which first predicts distance fields and then optimizes 3D humans and objects based on the distance fields.

All of the above optimization-based methods only rely on optimization targets~(\textit{e.g.}, 2D silhouettes) without considering image context.
One of the limitations of such an approach is vulnerability to imperfect optimization targets.
Since their optimization targets are acquired by estimation, the targets contain estimation errors, and some optimization targets are ambiguous~(\textit{e.g.}, depth ambiguity of 2D silhouettes) for reconstruction.
Accordingly, their optimization methods often fail by becoming biased toward the imperfect optimization targets.
Different from these methods, our CONTHO is an end-to-end learning approach that is free from the above issues.
This is because, in the inference stage, the system produces outputs based on the data-driven knowledge from the training data instead of being optimized towards imperfect targets.
Despite such a strength, we found that the learning-based systems can be vulnerable to being biased to specific contexts within an image, which we call an \emph{undesired human-object correlation}.
In this work, we unveil the undesired human-object correlation in the joint learning of 3D humans and objects and address it with our CRFormer.

\noindent\textbf{3D human reconstruction.}
Most of the 3D human reconstruction methods~\cite{kanazawa2018end, kolotouros2019convolutional, kolotouros2019learning,kocabas2020vibe,choi2021beyond,pymaf2021,kocabas2021pare,choi2022learning,li2022cliff, park2023extract, nam2023cyclic, choi2023rethinking, moon20223d} are based on parametric 3D human model (\textit{i.e.}, SMPL~\cite{loper2015smpl}).
HMR~\cite{kanazawa2018end} proposed an end-to-end learning framework that introduced adversarial loss to reconstruct a plausible 3D human mesh.
PARE~\cite{kocabas2021pare} used a part-guided attention network to ensure robustness in occlusions.
Hand4Whole~\cite{moon2022accurate} proposed a whole-body 3D human mesh estimation framework to reconstruct 3D human body, hand, and face with features from the 3D positional pose-guided pooling.
In our method, we bring the 3D body and hand reconstruction pipeline of Hand4Whole for initial 3D human mesh reconstruction.

\noindent\textbf{3D object reconstruction.}
One of the main approaches~\cite{tekin2018real,peng2019pvnet,xiang2018posecnn,wang2019densefusion,di2021so} of 3D object mesh reconstruction is to predict 6DoF pose (\textit{i.e.}, rotation and translation) of a given object mesh template after classifying the object category.
PoseCNN~\cite{xiang2018posecnn} is a pioneering work that proposes a convolutional neural network for object pose estimation.
SO-Pose~\cite{di2021so} employs self-occlusion information to predict accurate object pose.
ZebraPose~\cite{su2022zebrapose} proposed a coarse-to-fine surface encoding technique for 6DoF object pose estimation.
Our CONTHO also estimates the 6DoF object pose as an initial prediction and refines it considering human-object contact.
\section{CONTHO}
\label{sec:contho}
Figure~\ref{fig:pipeline} shows the overall pipeline of our CONTHO, which consists of three stages: initial reconstruction, 3D-guided contact estimation, and contact-based refinement.

\subsection{Initial reconstruction}
Given concatenated inputs~$\mathbf{I}_{\text{input}} \in \mathbb{R}^{5 \times H \times W}$ of image~$\mathbf{I}$, human segmentation~$\mathbf{S}_{\text{h}}$, and object segmentation~$\mathbf{S}_{\text{o}}$, we obtain the initial 3D human and object meshes~($\mathbf{M}_{\text{h}} \in \mathbb{R}^{431 \times 3}$ and $\mathbf{M}_{\text{o}} \in \mathbb{R}^{64 \times 3}$), where $H$ and $W$ denote the height and width of the image, respectively. 
Following previous works~\cite{bhatnagar2022behave, xie2022chore}, human and object segmentations are obtained from DetectronV2~\cite{wu2019detectron2} for both training and inference.
From the inputs~$\mathbf{I}_{\text{input}}$, a backbone network~(\textit{i.e.}, ResNet-50~\cite{he2016deep}) extracts an image feature $\mathbf{F} \in \mathbb{R}^{2048 \times H/32 \times W/32}$.
To obtain the initial 3D human mesh $\mathbf{M}_{\text{h}}$, we predict human body parameters $\mathbf{\theta}_{\text{body}} \in \mathbb{R}^{76}$ and hand parameters $\mathbf{\theta}_{\text{hand}} \in \mathbb{R}^{90}$ of the SMPL+H model~\cite{loper2015smpl} from the image feature $\mathbf{F}$.
Then, the predicted parameters are forwarded to SMPL+H model to obtain a 3D human mesh.
To reduce computational burden, the obtained 3D human mesh is downsampled with a sampling algorithm~\cite{ranjan2018generating}.
To obtain initial 3D object mesh $\mathbf{M}_{\text{o}}$, we predict object rotation~$\mathbf{R}_{\text{o}}$ and translation~$\mathbf{t}_{\text{o}}$ from the image feature~$\mathbf{F}$, given a 3D object mesh template as in prior works~\cite{xie2022chore,xie2023visibility}.
The overall design of the initial reconstruction module follows a state-of-the-art whole-body 3D human mesh reconstruction method~\cite{moon2022accurate} with modifications to only predict the human body and hands.
We provide a detailed description of the architecture in the supplementary material.

\subsection{3D-guided contact estimation}
In this stage, \textcolor[RGB]{230,170,10}{\textbf{ContactFormer}} predicts human and object contact maps~($\mathbf{C}_{\text{h}} \in \mathbb{R}^{431}$ and $\mathbf{C}_{\text{o}} \in \mathbb{R}^{64}$) from 3D vertex features~($\mathbf{F}_{\text{vh}}$ and $\mathbf{F}_{\text{vo}}$) extracted based on initially reconstructed 3D human and object meshes~($\mathbf{M}_{\text{h}}$ and $\mathbf{M}_{\text{o}}$).

\noindent\textbf{3D vertex feature extraction.} 
3D vertex features ($\mathbf{F}_{\text{vh}}$ and $\mathbf{F}_{\text{vo}}$) consist of vertex-aligned features and per-vertex 3D coordinates.
The vertex-aligned features are obtained by grid sampling of the image feature $\mathbf{F}$ with ($x$, $y$) positions of the projected 3D vertices of initial meshes~($\mathbf{M}_{\text{h}}$ and $\mathbf{M}_{\text{o}}$) to image space.
After grid sampling, we apply a 1-by-1 convolution to the vertex-aligned features to reduce the channel dimension from $2048$ to $256$.
Subsequently, we obtain 3D vertex features~($\mathbf{F}_{\text{vh}}$ and $\mathbf{F}_{\text{vo}}$) by concatenating the vertex-aligned features and per-vertex 3D coordinates of the initial meshes ($\mathbf{M}_{\text{h}}$ and $\mathbf{M}_{\text{o}}$).
Therefore, the final dimensions of the 3D vertex features of the human and object are~$\mathbf{F}_{\text{vh}} \in \mathbb{R}^{(256+3) \times 431}$ and $\mathbf{F}_{\text{vo}} \in \mathbb{R}^{(256+3) \times 64}$.
The 3D vertex features contain rich contextual information around the 3D mesh vertices, allowing 3D guidance for human-object contact estimation.
The 3D vertex features are passed to ContactFormer, the contact estimation Transformer.

\noindent\textbf{Human-object contact estimation.}
Given the 3D vertex features~($\mathbf{F}_{\text{vh}}$ and $\mathbf{F}_{\text{vo}}$), ContactFormer predicts human and object contact maps~($\mathbf{C}_{\text{h}}$ and $\mathbf{C}_{\text{o}}$).
To encourage the ContactFormer to focus on relevant information across humans and objects, we perform a cross-attention operation between 3D vertex features of humans and objects with cross-attention (CA) Transformers~\cite{vaswani2017attention}.
Then, the contact maps ($\mathbf{C}_{\text{h}}$ and $\mathbf{C}_{\text{o}}$) are predicted with fully-connected (FC) layers, followed by a sigmoid activation function.

\subsection{Contact-based refinement}
In this stage, \textcolor[RGB]{0,130,245}{\textbf{CRFormer}} provides refined 3D human and object meshes ($\mathbf{M}^{\ast}_{\text{h}}$ and $\mathbf{M}^{\ast}_{\text{o}}$) from the 3D vertex features ($\mathbf{F}_{\text{vh}}$ and $\mathbf{F}_{\text{vo}}$) and the contact maps ($\mathbf{C}_{\text{h}}$ and $\mathbf{C}_{\text{o}}$).

\noindent\textbf{Contact-based masking.}
Based on the contact maps ($\mathbf{C}_{\text{h}}$ and $\mathbf{C}_{\text{o}}$), we mask a part of 3D vertex features ($\mathbf{F}_{\text{vh}}$ and $\mathbf{F}_{\text{vo}}$) that are not in contact with zero vectors, to remain only features corresponding to human-object contact.
We denote the masked 3D vertex features of the human and object with $\mathbf{F}^{\text{c}}_{\text{vh}}$ and $\mathbf{F}^{\text{c}}_{\text{vo}}$, respectively.
This contact-based masking feature aggregation technique is our core strategy to force the contact maps to be the main signal for CRFormer to indicate which features to focus on.
Additionally, by removing features from non-contacting parts that are unrelated to human-object contact, we prevent learning undesired human-object correlation that is detrimental to accurate refinement.
We provide further discussion about the effectiveness of contact-based masking in Section~\ref{sec:ablation_study}.

\noindent\textbf{3D human and object refinement.}
The masked 3D vertex features ($\mathbf{F}^{\text{c}}_{\text{vh}}$ and $\mathbf{F}^{\text{c}}_{\text{vo}}$) and original 3D vertex features ($\mathbf{F}_{\text{vh}}$ and $\mathbf{F}_{\text{vo}}$) are processed with a combination of cross-attention (CA) and self-attention (SA) Transformers~\cite{vaswani2017attention}, to obtain refined 3D human and object meshes ($\mathbf{M}_{\text{h}}^{\text{*}}$ and $\mathbf{M}_{\text{o}}^{\text{*}}$).
$\mathbf{F}^{\text{c}}_{\text{vh}}$ and $\mathbf{F}^{\text{c}}_{\text{vo}}$ are passed to CA Transformers to process relevant information across human and object.
As $\mathbf{F}^{\text{c}}_{\text{vh}}$ and $\mathbf{F}^{\text{c}}_{\text{vo}}$ only contain features in contact, the CA Transformers mainly process contextual information related to human-object contact.
$\mathbf{F}_{\text{vh}}$ and $\mathbf{F}_{\text{vo}}$ are passed separately to SA Transformers to infer each own 3D positional information, without considering human-object interaction.
The SA Transformers mainly process contextual information related to non-contacting parts of the human and object.
This combination of CA and SA transformers prevents excessive bias or disregard of the contact information.
Lastly, the outputs of CA and SA Transformers are added and processed with additional SA Transformers followed by FC layers to produce refined 3D human and object meshes ($\mathbf{M}^{\ast}_{\text{h}}$ and $\mathbf{M}^{\ast}_{\text{o}}$).

\subsection{Loss functions}
Our proposed CONTHO is trained in an end-to-end manner by minimizing loss function $L$, defined as follows:
\begin{equation}
\begin{split}
L &= L_{\text{contact}} + L_{\text{refine}} + L_{\text{init}},
\end{split}
\end{equation}
where $L_{\text{contact}}$ is a binary-cross entropy loss between predicted and GT contact maps~($\mathbf{C}_{\text{h}}$ and $\mathbf{C}_{\text{o}}$).
The $L_{\text{refine}}$ is defined as
\begin{equation}
\begin{split}
L_{\text{refine}} &= L_{\text{vertex}} + L_{\text{edge}},
\end{split}
\end{equation}
where $L_{\text{vertex}}$ is a L1 distance between predicted and GT per-vertex 3D coordinates of refined human and object meshes ($\mathbf{M}^{\ast}_{\text{h}}$ and $\mathbf{M}^{\ast}_{\text{o}}$), and $L_{\text{edge}}$ is edge length consistency loss between predicted and GT edges of the refined human meshes ($\mathbf{M}^{\ast}_{\text{h}}$).
The $L_{\text{init}}$ is defined as
\begin{equation}
\begin{split}
L_{\text{init}} &= L_{\text{param}} + L_{\text{coord}} + L_{\text{hbox}},
\end{split}
\end{equation}
where~$L_{\text{param}}$ is a L1 distance between predicted and GT SMPL+H parameters ($\theta_{\text{body}}$ and $\theta_{\text{hand}}$), 3D object rotation $\mathbf{R}_{\text{o}}$, and 3D object translation $\mathbf{t}_{\text{o}}$. 
$L_{\text{coord}}$ is a L1 distance between the predicted and GT human joint coordinates, consisting of 3D and 2D joint coordinates.
$L_{\text{hbox}}$ is a L1 distance between the predicted and GT bounding boxes of the hands.
We design $L_{\text{init}}$ by modifying the loss function of Hand4Whole~\cite{moon2022accurate}.
For a detailed explanation, please refer to the supplementary material.
\section{Implementation details}
PyTorch~\cite{paszke2017automatic} is used for implementation. 
The backbone is initialized with pre-trained weights of publicly released Hand4Whole~\cite{moon2022accurate}. 
The weights are updated by Adam optimizer~\cite{kingma2014adam} with a mini-batch size of 16. 
The region of the reconstruction target is cropped using a GT box in both the training and the testing stages following previous works~\cite{xie2022chore, xie2023visibility}.
Data augmentations, including scaling, rotation, and color jittering, are performed in training.
The initial learning rate is set to $10^{-4}$ and reduced by a factor of 10 after the 30\textit{th} epoch.
We train the model for 50 epochs with an NVIDIA RTX 2080 Ti GPU. 
\section{Experiments}
\label{sec:experiments}
\subsection{Datasets}
BEHAVE~\cite{bhatnagar2022behave, xie2024rhobin} and InterCap~\cite{huang2022intercap} datasets are used for our experiments.
BEHAVE~\cite{bhatnagar2022behave, xie2024rhobin} is a dataset that captures the interactions of 8 human subjects and 20 objects.
We follow CHORE~\cite{xie2022chore} for the split of BEHAVE~\cite{bhatnagar2022behave, xie2024rhobin} for a fair comparison.
InterCap~\cite{huang2022intercap} is another human-object interaction dataset containing 10 human subjects with 10 objects.
Following the prior work~\cite{xie2023visibility}, we split the dataset accordingly.
For both datasets, we labeled contact maps on 3D human and object vertices with a 3D distance threshold of 5$cm$ between human and object.

\subsection{Evaluation metrics}
\noindent\textbf{Precision \& recall for contact estimation ($\text{Contact}^{\text{est}}_{\text{p}}$, $\text{Contact}^{\text{est}}_{\text{r}}$).}
We evaluate human-object contact estimation with standard detection metrics: precision ($\text{Contact}^{\text{est}}_{\text{p}}$) and recall ($\text{Contact}^{\text{est}}_{\text{r}}$), following Huang~\etal~\cite{huang2022capturing}.
Unlike our CONTHO, which estimates both human and object contact maps, previous contact estimation methods~\cite{huang2022capturing,tripathi2023deco} only estimate human contact maps.
Thus, for comparison, we report evaluations on human contact maps for all methods.

\noindent\textbf{Chamfer distance~($\text{CD}_{\text{human}}$, $\text{CD}_{\text{object}}$).}
We evaluate 3D human and object reconstruction using Chamfer distance between predicted and GT meshes, following previous works of 3D human and object reconstruction~\cite{xie2022chore,xie2023visibility}.
Specifically, given the predicted 3D human and object meshes, we apply Procrustes alignment on combined 3D human and object meshes with the GT 3D human and object meshes.
With the aligned 3D human and object meshes, we measure the Chamfer distance from GT separately on 3D human (CD$_{\text{human}}$) and 3D object (CD$_{\text{object}}$), in centimeters.

\noindent\textbf{Precision \& recall for contact from reconstruction ($\text{Contact}^{\text{rec}}_{\text{p}}$, $\text{Contact}^{\text{rec}}_{\text{r}}$).}
To evaluate 3D human and object reconstruction, especially in terms of contact, we further adopt standard detection metrics for the reconstructed 3D human and object meshes.
We obtain a contact map by classifying human vertices within 5$cm$ of the object mesh.
Then, we measure precision ($\text{Contact}^{\text{rec}}_{\text{p}}$) and recall ($\text{Contact}^{\text{rec}}_{\text{r}}$) between the human contact map and the GT counterpart.

\subsection{Ablation study}
\label{sec:ablation_study}
We carry out the ablation study by training and evaluating all methods on BEHAVE~\cite{bhatnagar2022behave}.

\noindent\textbf{Effectiveness of 3D-guided contact estimation.}
In Table~\ref{tab:abl_contactformer}, we show the effectiveness of 3D-guided contact estimation by examining the following variations: 1) variations of ContactFormer inputs and 2) variations of ContactFormer design.
The first block of Table~\ref{tab:abl_contactformer} shows that using 3D vertex features as the ContactFormer input largely outperforms other input variants.
The first variant uses a global average pooled (GAP) image feature.
The second and third variants follow existing methods~\cite{huang2022capturing,tripathi2023deco} by implementing their feature extractors into our framework.
Specifically, the second variant use a convolutional layer to extract per-vertex features from the image feature.
The third variant designs two encoders for human part and scene~\cite{cheng2021mask2former} and obtains features by applying cross-attention operation between two encoders' outputs.
One major difference of the second and third variants from ours is that they do not extract localized features based on 3D positions of 3D human and object.
Our 3D vertex feature contains localized contextual information around human and object regions by grid sampling on the image feature.
Additionally, the 3D positional information of 3D vertex features enables 3D geometric reasoning of human-object contact.
From these advantages, exploiting 3D vertex features outperforms other variants in contact estimation.
The second block of Table~\ref{tab:abl_contactformer} shows that our ContactFormer with CA Transformers achieves the best performance.
Compared to other designs, the cross-attention operation of the CA Transformers encourages ContactFormer to capture meaningful contextual information within the image.

\begin{table}[t]
\def\arraystretch{1.42}
\renewcommand{\tabcolsep}{0.7mm}
\footnotesize
\begin{center}
\scalebox{0.9}{
\begin{tabular}{>{\raggedright\arraybackslash}m{5.6cm}|>{\centering\arraybackslash}m{1.55cm}>  
{\centering\arraybackslash}m{1.55cm}}
\specialrule{.1em}{.05em}{0.0em}
Methods & \scalebox{0.92}{$\text{Contact}^{\text{est}}_{\text{p}}{\uparrow}$} & \scalebox{0.92}{$\text{Contact}^{\text{est}}_{\text{r}}{\uparrow}$} \\ \hline
\textbf{* Variations of ContactFormer input} \\
GAP feature & 0.645 & 0.481 \\
Per-vertex image feature~\cite{huang2022capturing} & 0.716 & 0.539 \\
Part-scene image feature~\cite{tripathi2023deco} & 0.719 & 0.556 \\
\textbf{3D vertex feature (Ours)} \vspace{0.4em} & \textbf{0.754} & \textbf{0.587} \\
\hline
\textbf{* Variations of ContactFormer design} \\
FC layers & 0.639 & 0.471 \\
SA Transformers & 0.725 & 0.575 \\
\textbf{ContactFormer (Ours)} & \textbf{0.754} & \textbf{0.587} \\
\specialrule{.1em}{-0.05em}{-0.05em}
\end{tabular}
}
\end{center}
\vspace*{-1.5em}
\caption{
\textbf{Ablation studies for 3D-guided contact estimation on BEHAVE~\cite{bhatnagar2022behave}.}
}
\vspace*{-1.0em}
\label{tab:abl_contactformer}
\end{table}

\begin{figure}[t]
\begin{center}
\includegraphics[width=1.0\linewidth]{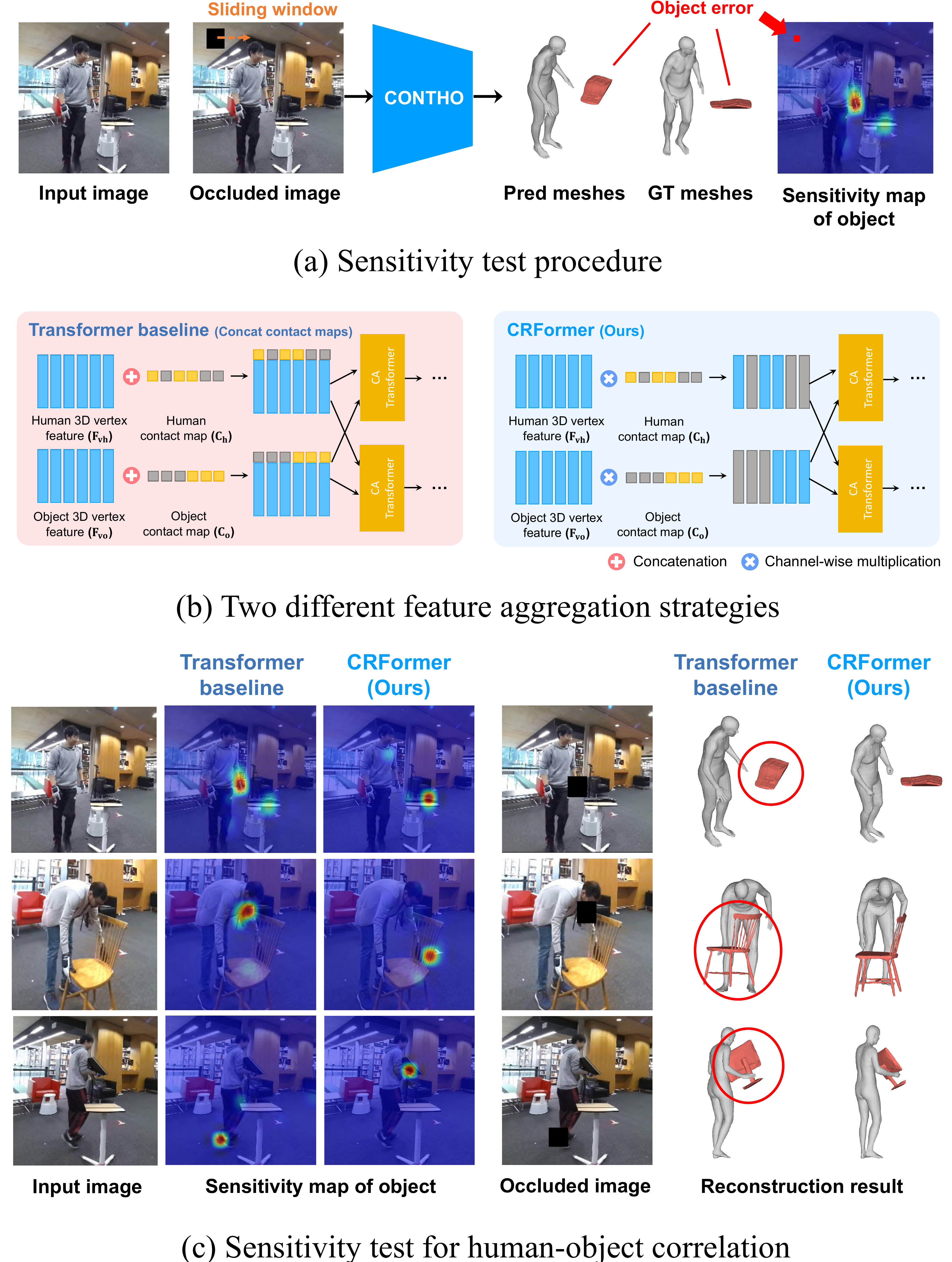}
\end{center}
\vspace*{-3.6mm}
\caption{
\textbf{Analysis of undesired human-object correlation on BEHAVE~\cite{bhatnagar2022behave}.}
We conduct a sensitivity test, inspecting which region is sensitive in reconstruction, for Transformer baseline and our CRFormer.
In the Transformer baseline, the object errors are sensitive to human regions not actually related to human-object interaction, as a result of undesired correlation.
In our CRFormer, the object errors are mostly sensitive around regions containing human-object contact.
}
\label{fig:correlation_analysis}
\vspace*{-1.8mm}
\end{figure}

\begin{figure*}[t]
\begin{center}
\includegraphics[width=1.0\linewidth]{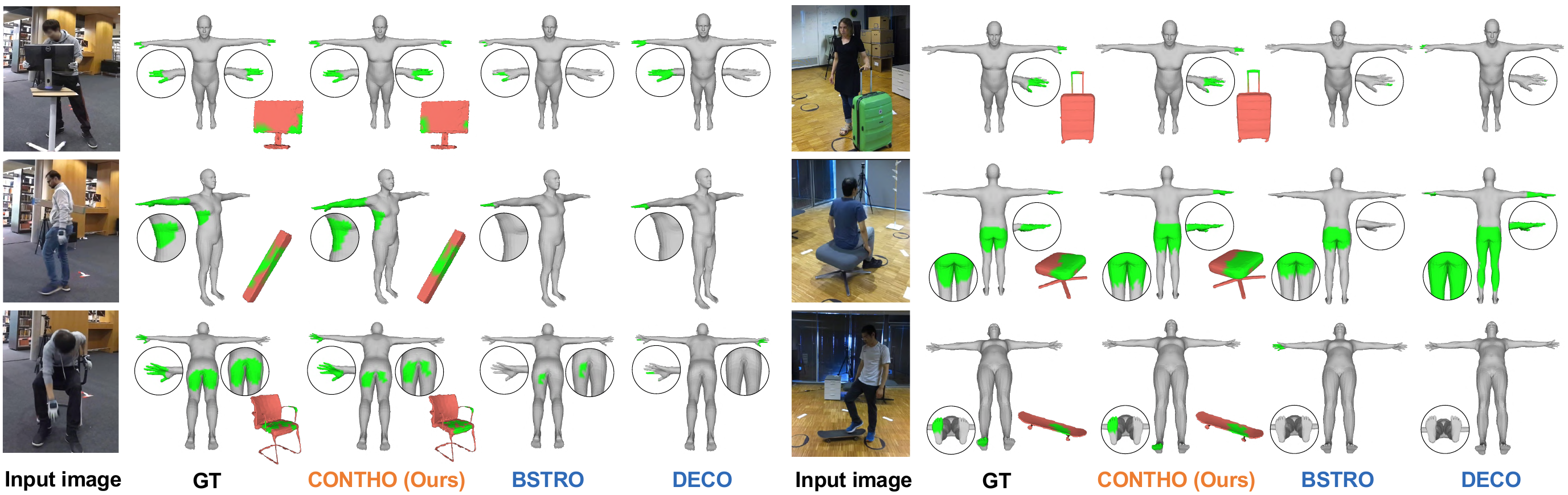}
\end{center}
\vspace*{-3.5mm}
\caption{
\textbf{Qualitative comparison of human-object contact estimation with BSTRO~\cite{huang2022capturing} and DECO~\cite{tripathi2023deco}, on BEHAVE~\cite{bhatnagar2022behave} (left) and InterCap~\cite{huang2022intercap} (right).}
The green color indicates the contacting regions.
}
\label{fig:qual_contact_est}
\vspace*{-2.0mm}
\end{figure*}

\noindent\textbf{Effectiveness of contact-based refinement.}
In Table~\ref{tab:abl_crformer}, we justify our proposed contact-based refinement by conducting ablation studies as follows: 1) variations of feature aggregation and 2) variations of CRFormer design.
The first block of Table~\ref{tab:abl_crformer} validates the effectiveness of contact-based masking of CRFormer compared to other variants of feature aggregation strategies.
Comparing the first and other variants shows the impact of contact maps in refinement.
As the contact maps are strong cues for understanding human-object interactions, utilizing the contact maps significantly enhances refinement performance, especially in contact-related metrics. 
Among all variants, our proposed contact-based masking (the fourth variant) achieves the best performance, as the masking strategy explicitly highlights contact maps as a key signal for refinement, unlike other aggregation strategies.
Additionally, the contact-based masking prevents learning undesired human-object correlation by removing unnecessary features unrelated to human-object interaction.
Due to such reasons, our proposed contact-based masking outperforms other aggregation strategies by significant margins.
The second block of Table~\ref{tab:abl_crformer} validates our CRFormer design as a combination of CA Transformers and SA Transformers.
The CA Transformers have strength in encouraging attention to the relevant information across human and object.
Differently, the SA Transformer has strength in learning positional relationships between human and object, separately.
By combining the advantages of each Transformer, our CRFormer achieves the best performance.

\begin{table}[t]
\def\arraystretch{1.42}
\renewcommand{\tabcolsep}{0.8mm}
\footnotesize
\begin{center}
\scalebox{0.9}{
\begin{tabular}{>{\raggedright\arraybackslash}m{4.2cm}|>{\centering\arraybackslash}m{1.05cm}>{\centering\arraybackslash}m{0.9cm}>{\centering\arraybackslash}m{1.15cm}>{\centering\arraybackslash}m{1.15cm}}
\specialrule{.1em}{.05em}{0.0em}
Methods & \scalebox{0.88}{$\text{CD}_{\text{human}}{\downarrow}$} & \scalebox{0.88}{$\text{CD}_{\text{object}}{\downarrow}$}  & 
\scalebox{0.88}{$\text{Contact}^{\text{rec}}_{\text{p}}{\uparrow}$} & \scalebox{0.88}{$\text{Contact}^{\text{rec}}_{\text{r}}{\uparrow}$} \\ \hline
Initial reconstruction & 5.70 & 10.86 & 0.547 & 0.394 \\
\hline
\textbf{* Variations of feature aggregation} \\
Without contact maps & 5.13 & 8.77 &  0.601 & 0.456 \\
Add contact maps & 5.12 & 8.54 & 0.616 & 0.483 \\
Concat contact maps & 5.18 & 8.65 & 0.618 & 0.477 \\
\textbf{Contact-based masking (Ours)}& \textbf{4.99} & \textbf{8.42} & \textbf{0.628} & \textbf{0.496} \\
\hline
\textbf{* Variations of CRFormer design} \\
CRFormer w/o CA Transformers & 5.40 & 9.03 & 0.591 & 0.419 \\
CRFormer w/o SA Transformers & 5.49 & 8.88 & 0.598  & 0.473 \\
\textbf{CRFormer (Ours)} & \textbf{4.99} & \textbf{8.42} & \textbf{0.628} & \textbf{0.496} \\
\specialrule{.1em}{-0.05em}{-0.05em}
\end{tabular}
}
\end{center}
\vspace*{-1.2em}
\caption{
\textbf{Ablation studies for contact-based refinement on BEHAVE~\cite{bhatnagar2022behave}.}
}
\vspace*{-1.0em}
\label{tab:abl_crformer}
\end{table}

\noindent\textbf{Analysis of undesired human-object correlation.}
In this section, we provide an in-depth analysis of undesired human-object correlation, which is detrimental to plausible reconstruction.
Human-object correlation is beneficial for learning 3D human and object reconstruction, in most cases.
However, the reconstruction network can be biased by the strong correlation between human and object poses and marginalize evidence within the image.
In the case of Figure~\ref{fig:undesired_bias}, 3D pose of the object~(\textit{i.e.}, monitor) is primarily determined by the human head, ignoring image evidence.
Although such an undesired correlation is detrimental to plausible reconstruction, there has not been much discussion for 3D human and object reconstruction. 
Consequently, we analyze the undesired human-object correlation issue using a sensitivity test, motivated by pioneering works~\cite{zeiler2014visualizing, kocabas2021pare}.
Figure~\ref{fig:correlation_analysis}~(a) shows the procedure of the sensitivity test.
Given an input image, we create an occluded image with an occluding patch for each pixel of the input image over a sliding window.
Then, we measure object reconstruction error~(\textit{i.e.,} CD$_{\text{object}}$) from each occluded image.
Repeating this process for all image regions yields a sensitivity map that indicates which regions in an image the object error is correlated with.

\begin{table}[t]
\def\arraystretch{1.42}
\renewcommand{\tabcolsep}{0.8mm}
\footnotesize
\begin{center}
\scalebox{0.9}{
\begin{tabular}{p{1.5cm}|>{\centering\arraybackslash}m{3.8cm}|>{\centering\arraybackslash}m{1.6cm}>{\centering\arraybackslash}m{1.6cm}}
\specialrule{.1em}{.05em}{0.0em}
Datasets & Methods & \scalebox{0.92}{$\text{Contact}^{\text{est}}_{\text{p}}{\uparrow}$} & \scalebox{0.92}{$\text{Contact}^{\text{est}}_{\text{r}}{\uparrow}$} \\ \hline
\multirow{4}{*}{BEHAVE} & POSA~\cite{hassan2021populating} & 0.514 & 0.299 \\
 & BSTRO~\cite{huang2022capturing} & 0.615 & 0.527 \\
  & DECO~\cite{tripathi2023deco} & 0.638 & 0.337 \\
 & \textbf{CONTHO (Ours)} & \textbf{0.754} & \textbf{0.587}  \\ \hline
\multirow{4}{*}{InterCap} & POSA~\cite{hassan2021populating} & 0.561 & 0.333 \\
 & BSTRO~\cite{huang2022capturing} & 0.506 & 0.427 \\
  & DECO~\cite{tripathi2023deco} & 0.635 & 0.479 \\
 & \textbf{CONTHO (Ours)} & \textbf{0.660} & \textbf{0.612}  \\
\specialrule{.1em}{-0.05em}{-0.05em}
\end{tabular}
}
\end{center}
\vspace*{-1.2em}
\caption{
\textbf{Quantitative comparison of human-object contact estimation with state-of-the-art methods on BEHAVE~\cite{bhatnagar2022behave} and InterCap~\cite{huang2022intercap}.}
}
\label{tab:quant_contact}
\end{table}

\begin{figure*}[t]
\begin{center}
\includegraphics[width=1.0\linewidth]{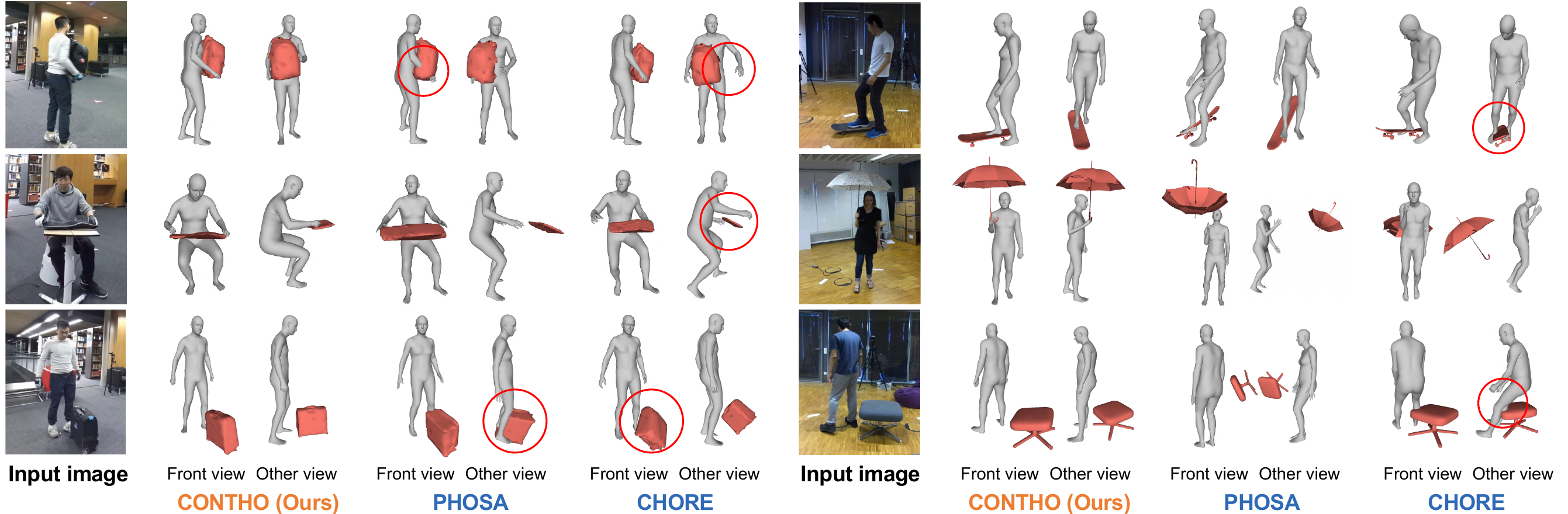}
\end{center}
\vspace*{-3.5mm}
\caption{
\textbf{Qualitative comparison of 3D human and object reconstruction with PHOSA~\cite{zhang2020perceiving} and CHORE~\cite{xie2022chore}, on BEHAVE~\cite{bhatnagar2022behave} (left) and InterCap~\cite{huang2022intercap} (right).}
We highlight their representative failure cases with red circles.
}
\label{fig:qual_reconstruction}
\vspace*{-1.8mm}
\end{figure*}

We conduct the sensitivity test for two different feature aggregation strategies of the contact-based refinement module, as shown in Figure~\ref{fig:correlation_analysis}~(b).
The Transformer baseline naively aggregates 3D vertex features and contact maps with concatenation.
As shown in Figure~\ref{fig:correlation_analysis}~(c), the sensitivity maps of the Transformer baseline are highly activated around human regions, which are not actually related to human-object interaction.
This means that 3D object reconstructions are much more correlated with human features from human regions than object features, although the human regions do not contain reasonable human-object interaction.
On the other hand, our CRFormer shows reasonable sensitivity maps, which are activated around object regions or human-object contacting regions.
This means that 3D object reconstruction is mainly correlated with the object's own regions or human-object contacting regions, which is a result of desirable correlation between human and object features.
Our CRFormer only considers human-object interaction among features from contacting regions through contact-based masking.
Thus, by explicitly preventing the exploitation of features unrelated to human-object contact, the CRFormer alleviates the undesired human-object correlation, producing accurate reconstruction results.
We provide more analysis examples in the supplementary material.

\subsection{Comparison with state-of-the-art methods}
We compare ours with previous state-of-the-art methods with two experimental protocols: 1) training \& evaluating all methods on BEHAVE~\cite{bhatnagar2022behave} and 2) training \& evaluating all methods on InterCap~\cite{huang2022intercap}.

\noindent\textbf{Human-object contact estimation.}
Figure~\ref{fig:qual_contact_est} and Table~\ref{tab:quant_contact} show that our CONTHO largely outperforms the state-of-the-art methods: POSA~\cite{hassan2021populating}, BSTRO~\cite{huang2022capturing}, and DECO~\cite{tripathi2023deco}.
BSTRO~\cite{huang2022capturing} and DECO~\cite{tripathi2023deco} often fail to capture human-object contact, especially in relatively small human parts~(\textit{e.g.,} hands), as human-object contact exists in a small area of the image.
In such difficult scenarios, our CONTHO is superior in capturing the local human-object contact, with the proposed 3D-guided contact estimation.
Under the 3D-guided contact estimation, 3D vertex feature provides explicit guidance on where to focus in local image regions, which allows the model to capture local context of human-object contact.
Furthermore, whereas previous methods only estimate human contact map, CONTHO additionally estimates object contact map along with human contact map.
This provides richer information about human-object interaction, showing which object parts are in contact with a human.

\begin{table}[t]
\def\arraystretch{1.65}
\renewcommand{\tabcolsep}{0.8mm}
\footnotesize
\begin{center}
\scalebox{0.9}{
\begin{tabular}{p{1.2cm}|>{\centering\arraybackslash}m{2.3cm}|>{\centering\arraybackslash}m{1.15cm}>{\centering\arraybackslash}m{0.95cm}>{\centering\arraybackslash}m{1.3cm}>{\centering\arraybackslash}m{1.3cm}}
\specialrule{.1em}{.05em}{0.0em}
Datasets & \multicolumn{1}{c|}{Methods} & \multicolumn{1}{c}{\scalebox{0.9}{$\text{CD}_{\text{human}}{\downarrow}$}} & \multicolumn{1}{c}{\scalebox{0.9}{$\text{CD}_{\text{object}}{\downarrow}$}}  & 
\multicolumn{1}{c}{\scalebox{0.9}{$\text{Contact}^{\text{rec}}_{\text{p}}{\uparrow}$}} & \multicolumn{1}{c}{\scalebox{0.9}{$\text{Contact}^{\text{rec}}_{\text{r}}{\uparrow}$}} \\ \hline
\multirow{3.25}{*}{BEHAVE} & PHOSA~\cite{zhang2020perceiving} & 12.17 & 26.62 & 0.393 & 0.266 \\
 & CHORE~\cite{xie2022chore} & 5.58 & 10.66 & 0.587 & 0.472 \\
 & \textbf{CONTHO (Ours)} \vspace{0.7mm} & \textbf{4.99} & \textbf{8.42} & \textbf{0.628} & \textbf{0.496} \\ \hline
 \multirow{3.25}{*}{InterCap} & PHOSA~\cite{zhang2020perceiving} & 11.20 & 20.57 & 0.228 & 0.159  \\ 
 & CHORE~\cite{xie2022chore} & 7.01 & 12.81 & 0.339 & 0.253 \\
 & \textbf{CONTHO (Ours)} \vspace{0.7mm} & \textbf{5.96} & \textbf{9.50} & \textbf{0.661}  & \textbf{0.432} \\
\specialrule{.1em}{-0.05em}{-0.05em}
\end{tabular}
}
\end{center}
\vspace*{-1.2em}
\caption{
\textbf{Quantitative comparison of 3D human and object reconstruction with state-of-the-art methods on BEHAVE~\cite{bhatnagar2022behave} and InterCap~\cite{huang2022intercap}.} 
}
\vspace*{-0.7em}
\label{tab:quant_cd}
\end{table}

\noindent\textbf{3D human and object reconstruction.}
Figure~\ref{fig:qual_reconstruction} and Table~\ref{tab:quant_cd} show that our CONTHO produces much better reconstruction results than the state-of-the-art methods: PHOSA~\cite{zhang2020perceiving} and CHORE~\cite{xie2022chore}.
PHOSA~\cite{zhang2020perceiving} and CHORE~\cite{xie2022chore} produce implausible reconstruction results, especially in terms of incorrect 3D object pose and human-object penetration.
The previous methods also fail when human and object are not in contact~(the last row in Figure~\ref{fig:qual_reconstruction}), producing reconstructions with an excessively short human-object distance.
This is largely due to their high reliance on imperfect optimization targets~(\textit{e.g.,} 2D silhouettes) during their optimization process.
On the other hand, our CONTHO accurately reconstructs 3D human and object meshes in both contacting and non-contacting cases by the following reasons.
First, our method reconstructs 3D human and object based on data-driven knowledge, instead of being optimized towards imperfect targets.
Second, our CRFormer learns human-object correlation mainly based on contact maps; focusing on contact regions in contacting cases, while learning no correlation in non-contacting cases.
As a consequence, our proposed method outperforms the previous reconstruction methods by a noticeable margin.

\section{Conclusion}
We propose CONTHO, a novel and powerful contact-based 3D human and object reconstruction method that utilizes human-object contact as the main driving signal in reconstruction.
For both accurate contact estimation and 3D human and object reconstruction, we propose a 3D-guided contact estimation pipeline and a contact-based refinement Transformer.
As a result, our CONTHO significantly outperforms previous methods in both human-object contact estimation and 3D human and object reconstruction.

\vspace*{+0.7em}
\noindent\textbf{Acknowledgements.}
This work was supported in part by the IITP grants [No. 2021-0-01343, Artificial Intelligence Graduate School Program (Seoul National University), No.2021-0-02068, and No.2023-0-00156], the NRF grant [No.2021M3A9E4080782] funded by the Korean government (MSIT).

\clearpage
\setcounter{page}{1}
\maketitlesupplementary

\setcounter{section}{0}
\setcounter{table}{0}
\setcounter{figure}{0}
\renewcommand{\thesection}{S\arabic{section}}   
\renewcommand{\thetable}{S\arabic{table}}   
\renewcommand{\thefigure}{S\arabic{figure}}

In this supplementary material, we present additional technical details and more experimental results that could not be included in the main manuscript due to the lack of pages.
The contents are summarized below:
\begin{itemize}
\vspace{3.0mm}
\item \ref{sec:supple_chairs}. Comparison with CHAIRS
\item \ref{sec:supple_run_time}. Comparison of running time
\item \ref{sec:supple_obj_contact_map}. Evaluation on object contact map 
\item \ref{sec:supple_hand4whole}. Details of initial reconstruction 
\item \ref{sec:supple_correlation}. More examples of undesired correlation 
\item \ref{sec:supple_more_qual}. More qualitative results
\item \ref{sec:supple_limitation}. Limitations and future works
\end{itemize}

\section{Comparison with CHAIRS}
\label{sec:supple_chairs}
Table~\ref{tab:suppl_chairs} shows that our CONTHO mostly outperforms CHAIRS~\cite{jiang2023full}, a recently published 3D human and object reconstruction method.
There are two core differences in the evaluation protocol between CHAIRS~\cite{jiang2023full} and the other state-of-the-art methods~\cite{zhang2020perceiving, xie2022chore}.
First, CHAIRS~\cite{jiang2023full} only reports reconstruction scores in specific object classes, such as chairs, table, yoga ball, and suitcase.
Second, its evaluation does not perform the Procrustes alignment before measuring the Chamfer distance.
Consequently, for a fair comparison with CHAIRS, Table~\ref{tab:suppl_chairs} reports the performance of our method following the CHAIRS evaluation process.
Similarly to the other previous methods~\cite{zhang2020perceiving, xie2022chore}, the CHAIRS is also an optimization-based method, which first reconstructs an object voxel and optimizes the 3D object mesh template on the reconstructed object voxel.
The optimization process totally depends on the 3D object voxel reconstruction without considering image features during the optimization.
Biased on the optimization target, CHAIRS often fails when the initial reconstruction provides an imperfect or noisy object voxel.
In contrast, our learning-based approach produces outputs based on data-driven knowledge obtained during training rather than optimizing on imperfect targets.
Furthermore, CHAIRS does not predict or exploit human-object contact for joint reconstruction.
Unlike CHAIRS, our CONTHO utilizes human-object contact information as a key signal for reconstruction by estimating and exploiting human-object contact maps for 3D human and object joint reconstruction.

\section{Comparison of running time}
\label{sec:supple_run_time}
Table~\ref{tab:suppl_running_time} shows that our CONTHO takes the shortest computational time compared to previous 3D human and object reconstruction methods.
The running time is measured in the same environment with the Intel Xeon Gold 6248R CPU and RTX 2080 Ti GPU.
For all methods, we exclude the pre-processing stage of acquiring human and object silhouettes.
PHOSA~\cite{zhang2020perceiving} and CHORE~\cite{xie2022chore} demand extremely long times, as their optimization processes iteratively fit 3D meshes with more than 100 iterations.
On the other hand, our CONTHO takes much less time, requiring only a single feed-forward for the inference under the learning-based approach.
Thus, our CONTHO has a significant advantage in running time compared to previous methods~\cite{zhang2020perceiving, xie2022chore}.

\begin{table}[t]
\def\arraystretch{1.42}
\renewcommand{\tabcolsep}{0.7mm}
\footnotesize
\begin{center}
\scalebox{0.85}{
\begin{tabular}{>{\raggedright\arraybackslash}m{1.35cm}|>{\centering\arraybackslash}m{1.0cm}>{\centering\arraybackslash}m{0.85cm}>{\centering\arraybackslash}m{0.85cm}>{\centering\arraybackslash}m{0.85cm}>{\centering\arraybackslash}m{0.85cm}>{\centering\arraybackslash}m{0.85cm}>{\centering\arraybackslash}m{0.85cm}>{\centering\arraybackslash}m{0.85cm}}
\specialrule{.1em}{.05em}{0.0em}
Methods & \multicolumn{2}{c}{Chair} & \multicolumn{2}{c}{Table} & \multicolumn{2}{c}{Yogaball} & \multicolumn{2}{c}{Suitcase} \\
 & \multicolumn{1}{c}{\scalebox{0.8}{$\text{CD}_{\text{human}}{\downarrow}$}} & \multicolumn{1}{c}{\scalebox{0.8}{$\text{CD}_{\text{object}}{\downarrow}$}} & \multicolumn{1}{c}{\scalebox{0.8}{$\text{CD}_{\text{human}}{\downarrow}$}} & \multicolumn{1}{c}{\scalebox{0.8}{$\text{CD}_{\text{object}}{\downarrow}$}} & \multicolumn{1}{c}{\scalebox{0.8}{$\text{CD}_{\text{human}}{\downarrow}$}} & \multicolumn{1}{c}{\scalebox{0.8}{$\text{CD}_{\text{object}}{\downarrow}$}} & \multicolumn{1}{c}{\scalebox{0.8}{$\text{CD}_{\text{human}}{\downarrow}$}} & \multicolumn{1}{c}{\scalebox{0.8}{$\text{CD}_{\text{object}}{\downarrow}$}} \\ \hline

CHAIRS \cite{jiang2023full} & 13.77 & 12.73 & 11.53 & 15.22 & 10.82 & \textbf{9.88} & 9.53 & 15.84 \\
\textbf{CONTHO (Ours)} & \textbf{5.94} & \textbf{9.91} & \textbf{6.57} & \textbf{9.30} & \textbf{6.42} & 10.67 & \textbf{4.81} & \textbf{8.36} \\
\specialrule{.1em}{-0.05em}{-0.05em}
\end{tabular}
}
\end{center}
\vspace*{-1.0em}
\caption{
\textbf{Quantitative comparison of 3D human and object reconstruction with CHAIRS~\cite{jiang2023full}.}
}
\vspace*{-0.1em}
\label{tab:suppl_chairs}
\end{table}

\begin{table}[t]
\def\arraystretch{1.55}
\renewcommand{\tabcolsep}{0.7mm}
\footnotesize
\begin{center}
\scalebox{0.92}{
\begin{tabular}{>{\centering\arraybackslash}m{2.6cm}>{\centering\arraybackslash}m{2.6cm}>{\centering\arraybackslash}m{3.3cm}}
\specialrule{.1em}{.05em}{0.0em}
PHOSA~\cite{zhang2020perceiving} & CHORE~\cite{xie2022chore} & \textbf{CONTHO (Ours)} \\ \hline
165.30 & 312.20 & \textbf{0.077} \\
\specialrule{.1em}{-0.05em}{-0.05em}
\end{tabular}
}
\end{center}
\vspace*{-1.0em}
\caption{
\textbf{Running time comparison between different methods.}
The unit is seconds per frame.
}
\label{tab:suppl_running_time}
\end{table}

\section{Evaluation on object contact map}
\label{sec:supple_obj_contact_map}
Compared to the previous human-object contact estimation methods~\cite{huang2022capturing,tripathi2023deco}, which only estimate a human contact map, our CONTHO additionally estimates an object contact map. 
We evaluate CONTHO in object contact maps by using the same evaluation metrics of human contact maps~(\textit{i.e.,}~$\text{Contact}^{\text{est}}_{\text{p}}$ and~$\text{Contact}^{\text{est}}_{\text{r}}$), described in Section~\ref{sec:experiments}.
For the BEHAVE dataset~\cite{bhatnagar2022behave}, the precision and recall scores of object contact estimation are 0.530 and 0.217, respectively.
For the InterCap dataset~\cite{huang2022intercap}, the precision and recall scores are 0.627 and 0.323, respectively.
Although our qualitative results~(Figure~\ref{fig:qual_contact_est}) show plausible estimations about object contact maps, the scores of object contact maps are relatively lower than those of human contact maps.
This is due to the lack of considerations for 3D object symmetry from these evaluation metrics~(\textit{i.e.}, $\text{Contact}^{\text{est}}_{\text{p}}$ and $\text{Contact}^{\text{est}}_{\text{r}}$).
For example, when touching a ball with a sphere shape, any part of the ball can be the correct contact region.
However, the evaluation metrics are vertex-to-vertex comparisons between predictions and GTs, where the vertex-to-vertex pairs are fixed.
Thus, the evaluation metrics cannot consider the symmetry of 3D objects, which results in an underestimation of the performance of object contact maps.

\begin{figure}[t]
\begin{center}
\includegraphics[width=1.0\linewidth]{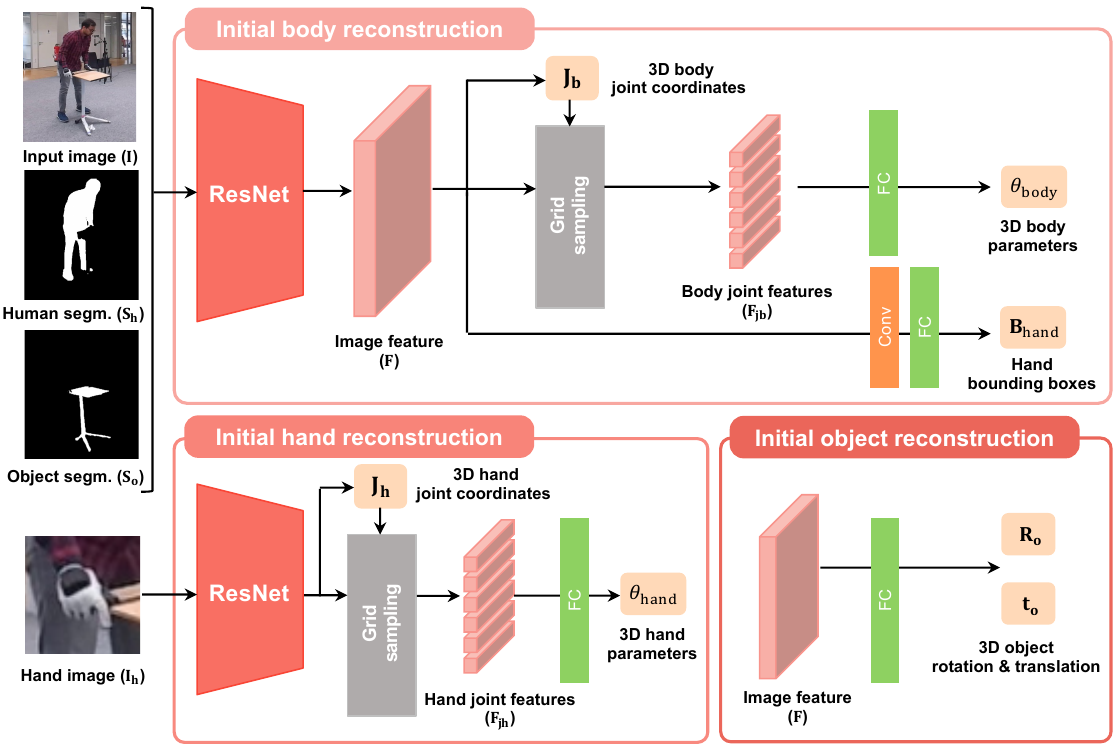}
\end{center}
\vspace*{-3.0mm}
\caption{
\textbf{The detailed architecture of the initial reconstruction stage in CONTHO.}
}
\label{fig:supple_hand4whole}
\vspace*{-1.0mm}
\end{figure}

\section{Details of initial reconstruction}
\label{sec:supple_hand4whole}
\subsection{Architectural design}
Our initial reconstruction follows Hand4Whole~\cite{moon2022accurate} with a few modifications.
Figure~\ref{fig:supple_hand4whole} specifically describes the backbone network in Figure~\ref{fig:pipeline}, which consists of two separate ResNets for reconstructing 1) 3D human body along with 3D object and 2) 3D human hands, respectively.

\noindent\textbf{Initial body reconstruction.}
To predict 3D body parameters~$\mathbf{\theta}_{\text{body}}$, we first extract an image feature~$\mathbf{F}$ with a backbone network~(\textit{i.e.}, ResNet-50~\cite{he2016deep}).
Then, we apply a 1-by-1 convolutional layer, followed by soft-argmax operation~\cite{sun2018integral}, to extract 3D human body joints~$\mathbf{J}_{\text{b}}$ from the image feature~$\mathbf{F}$ in a differentiable way.
The 3D human body joints~$\mathbf{J}_{\text{b}}$ are utilized to extract body joint features~$\mathbf{F}_{\text{jb}}$ by conducting grid sampling onto~($x$, $y$) positions of 3D body joint coordinates~$\mathbf{J}_{\text{b}}$ of the image feature~$\mathbf{F}$.
Based on the extracted body joint features~$\mathbf{F}_{\text{jb}}$, we predict 3D body parameters~$\mathbf{\theta}_{\text{body}}$ with a fully-connected layer after flattening the 3D joint features~$\mathbf{F}_{\text{jb}}$.
The image feature $\mathbf{F}$ is used further for the three subsequent processes.
First, the 2D hand bounding boxes~$\mathbf{B}_{\text{hand}}$ are obtained by forwarding the image feature to a combination of convolutional layers and fully-connected (FC) layers, where the 2D hand bounding boxes are passed to initial hand reconstruction.
Second, the image feature is also used in initial object reconstruction.
Third, we obtain the 3D vertex features, which are the core components of our CONTHO (Section~\ref{sec:contho}). 

\noindent\textbf{Initial hand reconstruction.}
Using the estimated 2D bounding boxes~$\mathbf{B}_{\text{hand}}$, we crop the input image~$\mathbf{I}$ to obtain hand image~$\mathbf{I}_{\text{h}}$, which serves as an input for initial hand reconstruction.
Similarly to the initial body reconstruction, our initial hand reconstruction acquires 3D hand parameters~$\mathbf{\theta}_{\text{hand}}$ with a similar process of extracting 3D hand joint coordinates~$\mathbf{J}_{\text{h}}$ and utilizing the 3D joint coordinates~$\mathbf{J}_{\text{h}}$ to obtain 3D joint features of hand~$\mathbf{F}_{\text{jh}}$, which will be passed to a fully-connected layer.
The 3D body parameters~$\mathbf{\theta}_{\text{body}}$ from initial body reconstruction and 3D hand parameters~$\mathbf{\theta}_{\text{hand}}$ from initial hand reconstruction are passed to SMPL+H model to get initial human mesh~$\mathbf{M}_{\text{h}}$.

\noindent\textbf{Initial object reconstruction.}
To predict 3D object parameters~($\mathbf{R}_{\text{o}}$ and $\mathbf{t}_{\text{o}}$), we process image feature~$\mathbf{F}$ from the initial body reconstruction and predict 3D object rotation~$\mathbf{R}_{\text{o}}$ and translation~$\mathbf{t}_{\text{o}}$ with a fully-connected layer.
In the end, the initial reconstruction stage obtains the initial object mesh~$\mathbf{M}_{\text{o}}$ along with the initial human mesh~$\mathbf{M}_{\text{h}}$, which are passed to the next stage of CONTHO, the 3D-guided contact estimation.

\subsection{Loss function design}
In Section~\ref{sec:contho}, the $L_{\text{init}}$ is designed to supervise the output of the initial reconstruction stage, with a few modifications of Hand4Whole~\cite{moon2022accurate}'s loss function.
The $L_{\text{init}}$ is defined as
\begin{equation}
\begin{split}
L_{\text{init}} &= L_{\text{param}} + L_{\text{coord}} + L_{\text{hbox}}.
\end{split}
\end{equation}

\noindent\textbf{Parameter loss~($L_{\text{param}}$).}
We minimize L1 loss between the predicted and GT parameters for 3D human body mesh~$\mathbf{\theta}_{\text{body}}$, 3D human hand mesh~$\mathbf{\theta}_{\text{hand}}$, and 3D object mesh~($\mathbf{R}_{\text{o}}$, and $\mathbf{t}_{\text{o}}$).

\noindent\textbf{Coordinate loss~($L_{\text{coord}}$).}
We utilize L1 loss between the predicted and GT human joint coordinates.
Specifically, the human joint coordinates consist of three types: 1) the extracted 3D body joint coordinates~$\mathbf{J}_{\text{b}}$ and 3D hand joint coordinates~$\mathbf{J}_{\text{h}}$, 2) 3D joint coordinates regressed from 3D human mesh~$\mathbf{M}_{\text{h}}$ with pre-defined regression matrix of SMPL+H, and 3) 2D joint coordinates, obtained by projecting the 3D joint coordinates from 3D human mesh~$\mathbf{M}_{\text{h}}$ to image space.

\noindent\textbf{Hand bounding boxes loss~($L_{\text{hbox}}$).}
We implement L1 loss between the predicted and GT bounding boxes of hand.
Specifically, the L1 distance for the center and scale of the bounding boxes are computed following Hand4Whole~\cite{moon2022accurate}.

\section{More examples of undesired correlation}
\label{sec:supple_correlation}
\begin{figure*}[t]
\begin{center}
\includegraphics[width=0.86\linewidth]{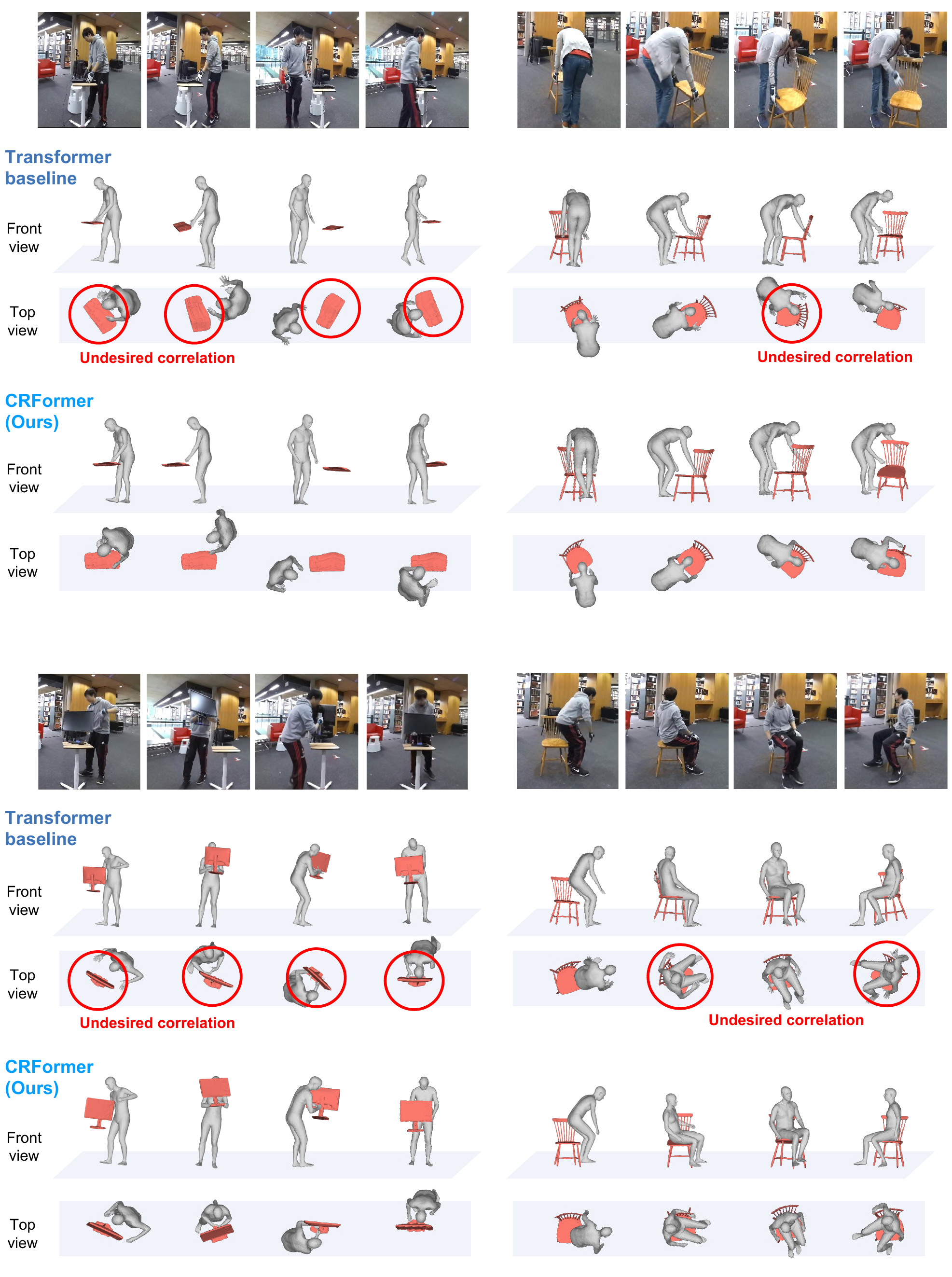}
\end{center}
\vspace*{-4.5mm}
\caption{
\textbf{Undesired correlation between human and object on BEHAVE~\cite{bhatnagar2022behave}.}
}
\label{fig:suppl_undesired_correlation_1}
\vspace*{-3.5mm}
\end{figure*}

In Figure~\ref{fig:suppl_undesired_correlation_1}, we provide more examples of the undesired correlation between human and object.
In these examples, we further show that the Transformer baseline with naive use of contact, learns undesired correlation between human and object.
As shown in Figure~\ref{fig:suppl_undesired_correlation_1}, the Transformer baseline outputs objects to face toward a certain human body part~(\textit{e.g.,} the back of a chair and a monitor display facing toward human face) or to be positioned in frequent interacting poses~(\textit{e.g.,} a keyboard in typing position and chair in regular sitting position). 
Unlike the baseline method, our CONTHO does not suffer from the undesired correlation with the novel contact-based refinement Transformer.

\begin{figure*}[t]
\begin{center}
\includegraphics[width=0.86\linewidth]{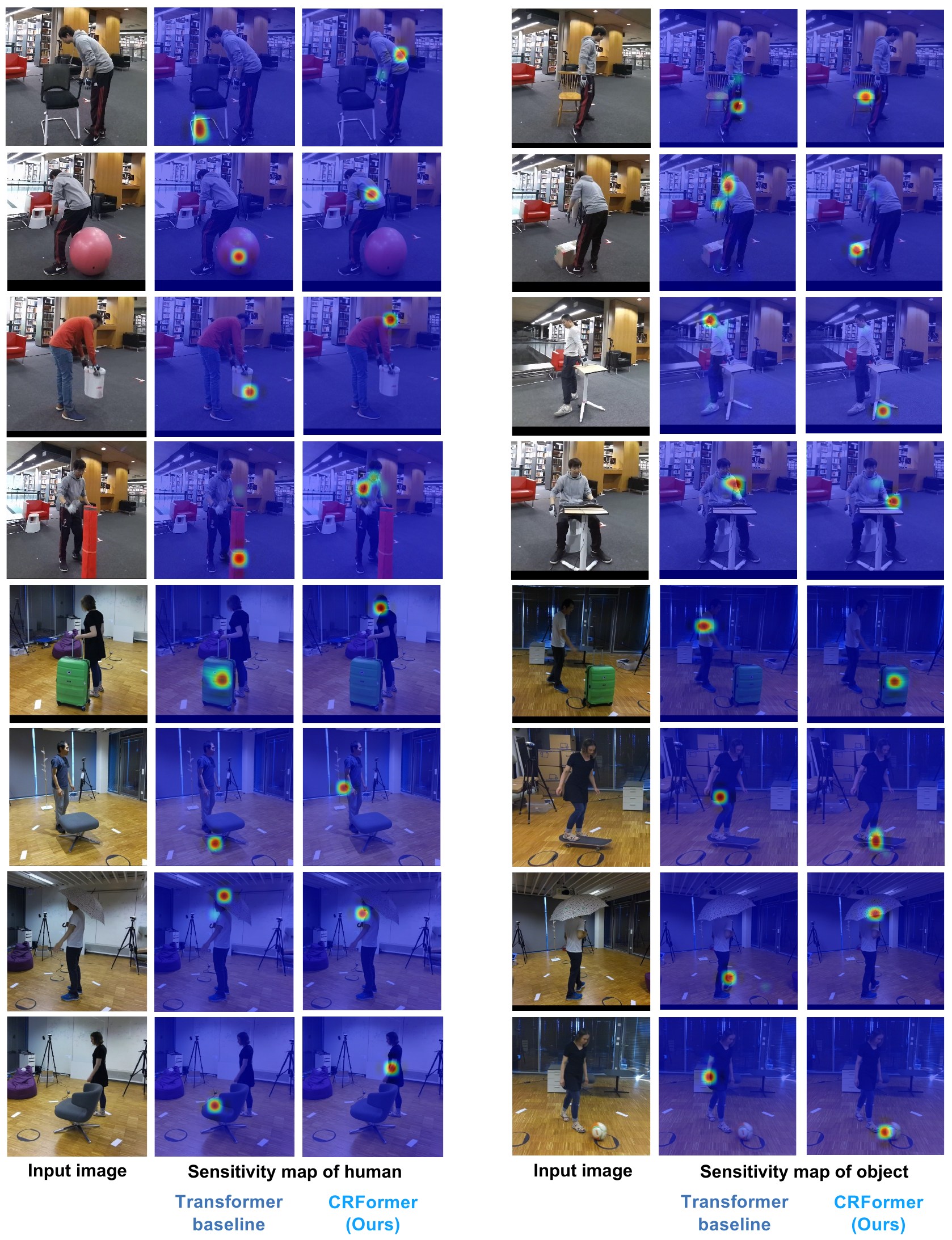}
\end{center}
\vspace*{-4.5mm}
\caption{
\textbf{Sensitivity tests of human errors (left) and object errors (right) on BEHAVE~\cite{bhatnagar2022behave} and InterCap~\cite{huang2022intercap}.}
}
\label{fig:suppl_sensitivity_test}
\vspace*{-3.5mm}
\end{figure*}

In Figure~\ref{fig:suppl_sensitivity_test}, we additionally show the undesired correlation between human and object with sensitivity tests.
In addition to the main manuscript, which only showed sensitivity tests of object errors, we show the results of our sensitivity test of human errors for the Transformer baseline and our CRFormer with more results for sensitivity tests of object errors on BEHAVE~\cite{bhatnagar2022behave} and InterCap~\cite{huang2022intercap}.
The results from sensitivity tests show that the Transformer baseline contains high sensitivity in the object regions for human errors and in human regions for object errors, indicating a high correlation between human and object.
Unlike the Transformer baseline, our CRFormer gives reasonable sensitivity of human regions for human errors and object regions for object errors on both datasets.

\section{More qualitative results}
\label{sec:supple_more_qual}
\begin{figure*}[t]
\begin{center}
\includegraphics[width=1.0\linewidth]{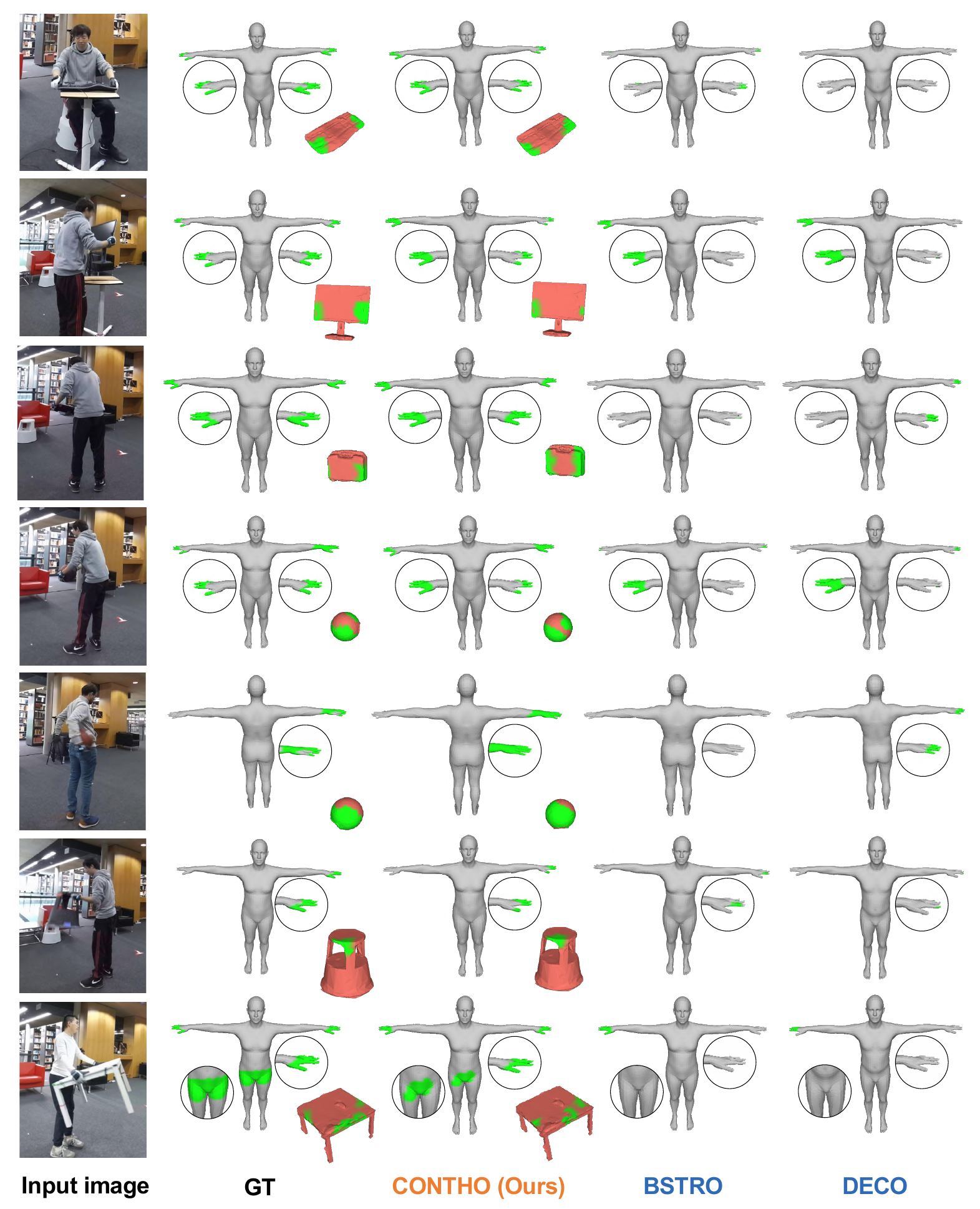}
\end{center}
\vspace*{-4.5mm}
\caption{
\textbf{Qualitative comparison of human-object contact estimation with BSTRO~\cite{huang2022capturing} and DECO~\cite{tripathi2023deco}, on BEHAVE~\cite{bhatnagar2022behave}.}
The green color indicates the contacting regions.
}
\label{fig:suppl_qual_contact_1}
\vspace*{-3.5mm}
\end{figure*}

\begin{figure*}[t]
\begin{center}
\includegraphics[width=1.0\linewidth]{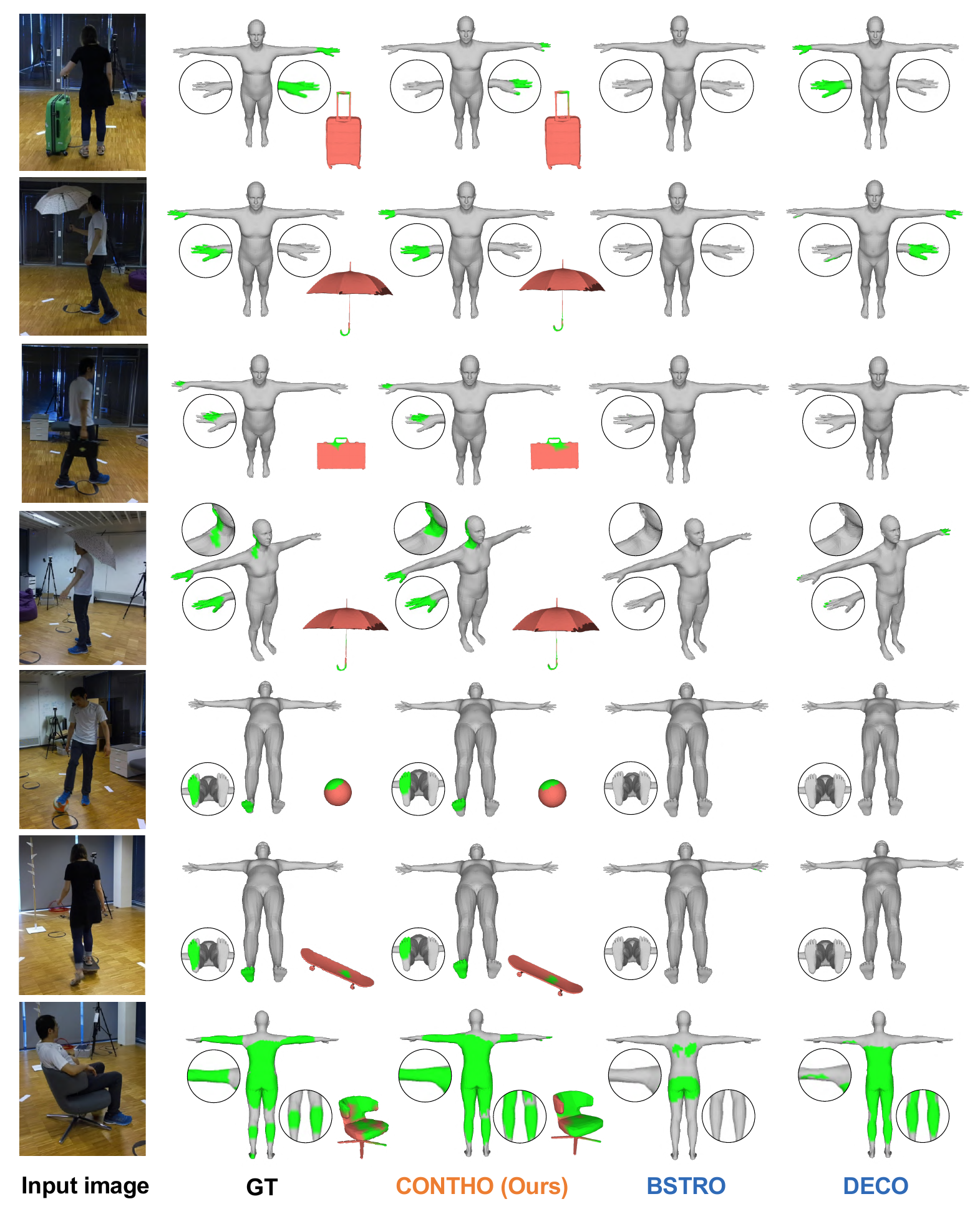}
\end{center}
\vspace*{-4.5mm}
\caption{
\textbf{Qualitative comparison of human-object contact estimation with BSTRO~\cite{huang2022capturing} and DECO~\cite{tripathi2023deco}, on InterCap~\cite{huang2022intercap}.}
The green color indicates the contacting regions.   
}
\label{fig:suppl_qual_contact_2}
\vspace*{-3.5mm}
\end{figure*}

\begin{figure*}[t]
\begin{center}
\includegraphics[width=1.0\linewidth]{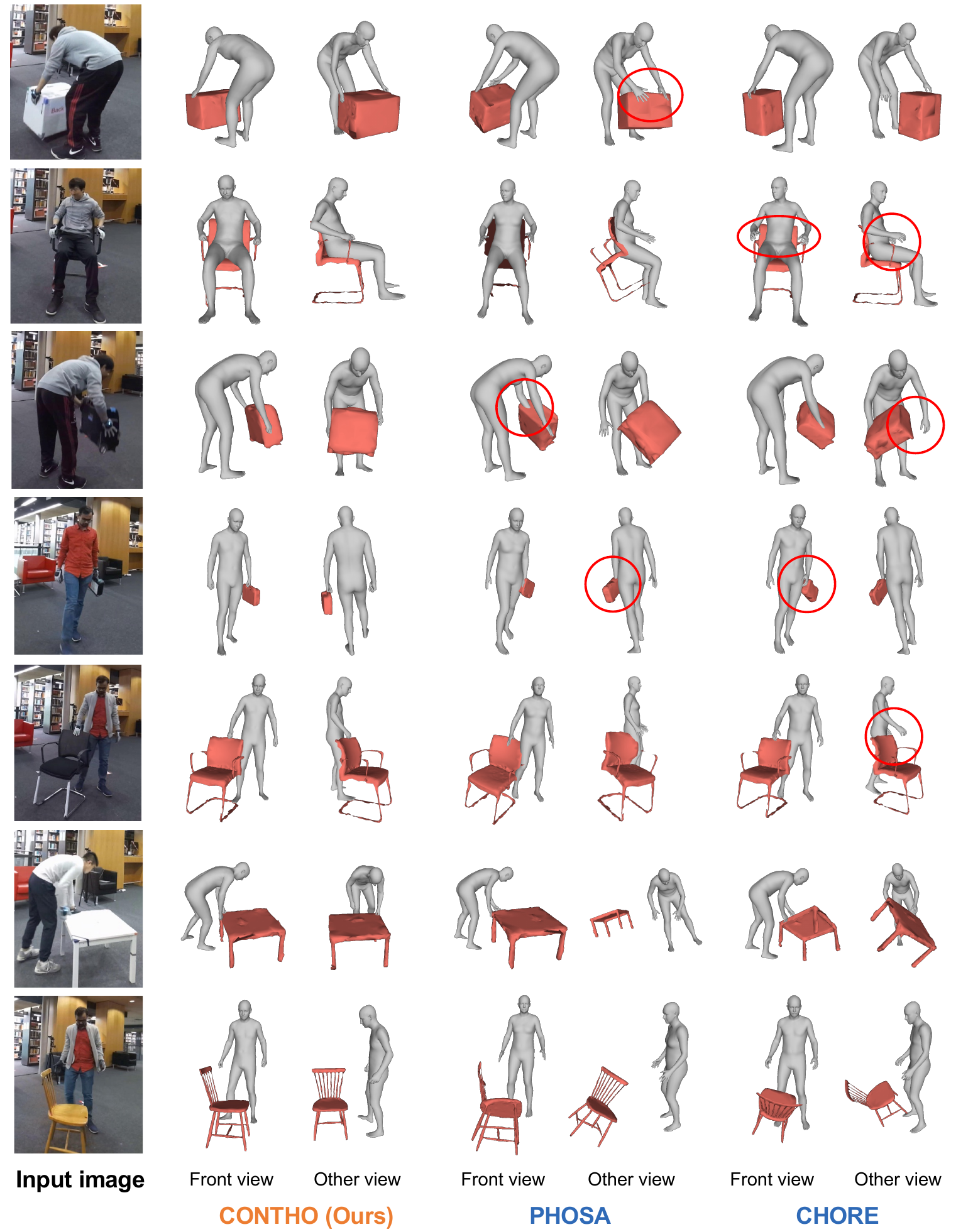}
\end{center}
\vspace*{-4.5mm}
\caption{
\textbf{Qualitative comparison of 3D human and object reconstruction with PHOSA~\cite{zhang2020perceiving} and CHORE~\cite{xie2022chore}, on BEHAVE~\cite{bhatnagar2022behave}.}
We highlight their representative failure cases with red circles.
}
\label{fig:suppl_qual_recon_1}
\vspace*{-3.5mm}
\end{figure*}

\begin{figure*}[t]
\begin{center}
\includegraphics[width=1.0\linewidth]{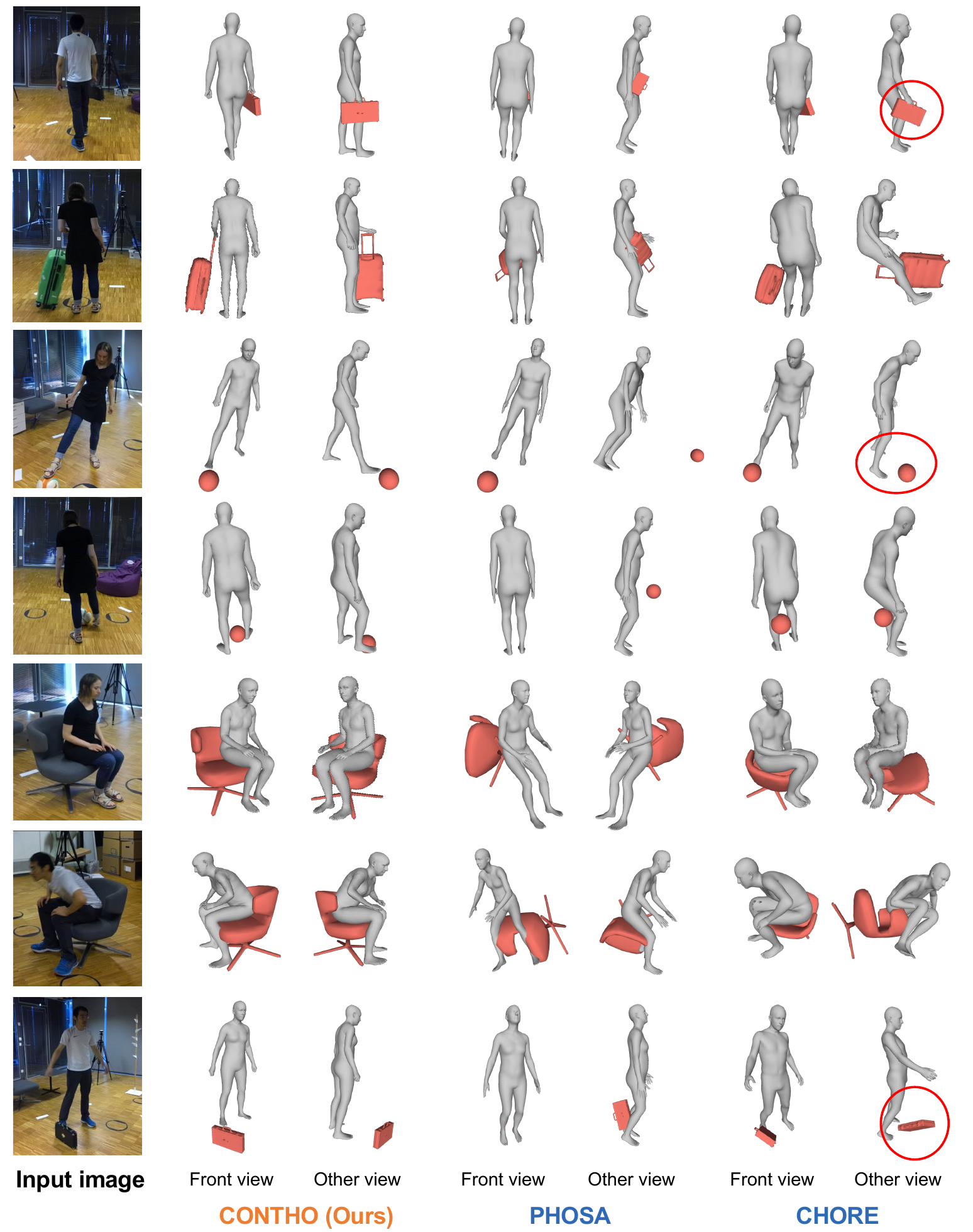}
\end{center}
\vspace*{-4.5mm}
\caption{
\textbf{Qualitative comparison of 3D human and object reconstruction with PHOSA~\cite{zhang2020perceiving} and CHORE~\cite{xie2022chore}, on InterCap~\cite{huang2022intercap}.}
We highlight their representative failure cases with red circles.
}
\label{fig:suppl_qual_recon_2}
\vspace*{-3.5mm}
\end{figure*}

We provide more qualitative comparisons of human-object contact estimation and 3D human and object reconstruction under two experimental protocols: 1) training \& evaluating all methods on BEHAVE~\cite{bhatnagar2022behave} and 2) training \& evaluating all methods on InterCap~\cite{huang2022intercap}.
Figure~\ref{fig:suppl_qual_contact_1} and Figure~\ref{fig:suppl_qual_contact_2} show that our CONTHO vastly outperforms previous contact estimation methods~(BSTRO~\cite{huang2022capturing} and DECO~\cite{tripathi2023deco}) on BEHAVE~\cite{bhatnagar2022behave} and InterCap~\cite{huang2022intercap}. 
Figure~\ref{fig:suppl_qual_recon_1} and Figure~\ref{fig:suppl_qual_recon_2} show that our CONTHO produces much accurate reconstruction results than previous reconstruction methods (PHOSA~\cite{zhang2020perceiving} and CHORE~\cite{xie2022chore}) on BEHAVE~\cite{bhatnagar2022behave} and InterCap~\cite{huang2022intercap}.

\begin{figure}[t]
\begin{center}
\includegraphics[width=1.0\linewidth]{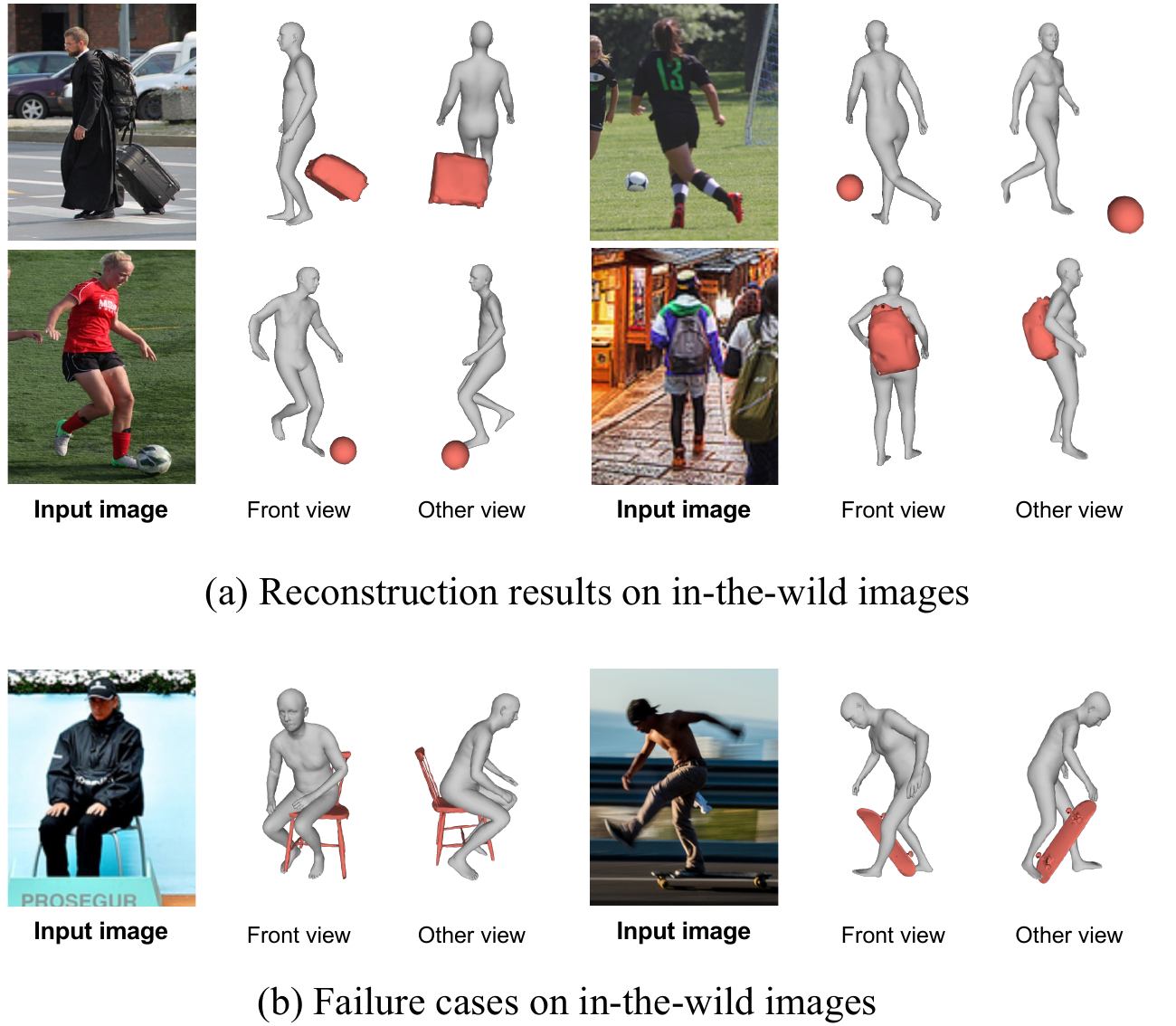}
\end{center}
\vspace*{-3.5mm}
\caption{
\textbf{Qualitative results of CONTHO on in-the-wild images.}
}
\label{fig:suppl_in_the_wild}
\vspace*{-2.5mm}
\end{figure}

\section{Limitations and future works}
\label{sec:supple_limitation}
\noindent\textbf{Generalization to in-the-wild images.}
Figure~\ref{fig:suppl_in_the_wild} shows qualitative reconstruction results of CONTHO on in-the-wild images of MSCOCO~\cite{lin2014mscoco} and MPII~\cite{andriluka2014mpii}.
To obtain the results, we used the network trained on BEHAVE~\cite{bhatnagar2022behave}, without any fine-tuning.
As shown in Figure~\ref{fig:suppl_in_the_wild} (b), our CONTHO fails on some in-the-wild images.
This is mainly due to domain gap between training datasets~\cite{bhatnagar2022behave} and in-the-wild images.
As the training datasets are acquired in restricted environments, the datasets contain much less diverse image appearances compared to in-the-wild datasets.
Due to such a domain gap problem, generalization on in-the-wild images is one of the challenges to be solved.

\noindent\textbf{Diversity of object shape.}
Our CONTHO covers a limited number of 3D object categories, included in the training datasets.
However, real-world objects are more diverse than the restrained categories of training objects.
Collecting more 3D object data in the real-world and learning interaction with the objects are crucial future research directions.

\noindent\textbf{Video as input.}
Our CONTHO aims to jointly reconstruct 3D human and object from a single image.
The recent emergence of a video-based method for 3D human and object reconstruction~\cite{xie2023visibility} suggests that reconstructing human and object from a video could be a promising direction.
Despite its strong performance, it utilizes future frames for the results of a certain frame, which we call an offline approach.
Instead, we think an online approach, which does not assume the availability of the future frame, could be closer to real-world applications, and we aim to extend our CONTHO for such an online approach.

\subsection*{License of the Used Assets}
\begin{itemize}
\vspace{1.0mm}
\item BEHAVE dataset~\cite{bhatnagar2022behave} is available for the sole purpose of performing non-commercial scientific research.
\item InterCap dataset~\cite{huang2022intercap} is released for non-commercial scientific research purposes.
\item BSTRO codes~\cite{huang2022capturing} are released under the MPI license.
\item DECO codes~\cite{tripathi2023deco} are released for non-commercial scientific research purposes.
\item PHOSA codes~\cite{zhang2020perceiving} are released under CC BY-NC 4.0.
\item CHORE codes~\cite{xie2022chore} is available for the sole purpose of performing non-commercial scientific research.
\end{itemize}

\clearpage

\clearpage

{\small
\bibliographystyle{ieee_fullname}
\bibliography{egbib}
}

\end{document}